%% file: iclr2024_conference.tex
\documentclass{article}

\PassOptionsToPackage{numbers}{natbib}
\usepackage{iclr2024_conference,times}

\usepackage[utf8]{inputenc} 
\usepackage[T1]{fontenc}    
\usepackage[colorlinks, linkcolor=black, urlcolor=black, citecolor=black]{hyperref}       
\usepackage{url}            
\usepackage{booktabs}       
\usepackage{amsfonts}       
\usepackage{nicefrac}       
\usepackage{microtype}      
\usepackage{xcolor}         
\usepackage{amsmath}
\usepackage{algpseudocode}
\usepackage{algorithm}
\usepackage{setspace}
\usepackage{tabularx}
\usepackage{multicol}
\usepackage{multirow}
\usepackage{graphicx}
\usepackage{tablefootnote}
\usepackage{titlesec}
\titlespacing{\paragraph}{\parindent}{0ex}{1ex}
\usepackage{listings}
\usepackage{xcolor}
\usepackage{subfig}
\usepackage{enumitem}
\usepackage{minitoc}
\usepackage{float}

\definecolor{codegreen}{rgb}{0,0.6,0}
\definecolor{codegray}{rgb}{0.5,0.5,0.5}
\definecolor{codepurple}{rgb}{0.58,0,0.82}
\definecolor{backcolour}{rgb}{0.95,0.95,0.95}

\lstdefinestyle{mystyle}{
    backgroundcolor=\color{backcolour},
    commentstyle=\color{codegreen},
    keywordstyle=\color{magenta},
    numberstyle=\tiny\color{codegray},
    stringstyle=\color{codepurple},
    basicstyle=\ttfamily\footnotesize,
    breakatwhitespace=false,         
    breaklines=true,                 
    captionpos=b,                    
    keepspaces=true,                 
    numbers=left,                    
    numbersep=5pt,                  
    showspaces=false,                
    showstringspaces=false,
    showtabs=false,                  
    tabsize=2
}
\iclrfinalcopy
\DeclareMathOperator*{\argmin}{argmin}

\title{GIO: Gradient Information Optimization for Training Dataset Selection}

\newcommand{\ourmethod}{Gradient Information Optimization}
\newcommand{\ourmethodabbrev}{\textsc{Gio}}
\newcommand{\secref}[1]{Section \ref{#1}}

\newcommand{\figref}[1]{Figure~\ref{#1}}
\newcommand{\Figref}[1]{Figure~\ref{#1}}
\newcommand{\tabref}[1]{Table~\ref{#1}}
\newcommand{\Tabref}[1]{Table~\ref{#1}}
\newcommand{\appref}[1]{Appendix \ref{#1}}

\newcommand{\appendixhead}%
{\begin{center}{\textbf{\huge Appendix}
\vspace{0.25in}}\end{center}}

\author{%
  Dante Everaert \\
  Amazon Search Science\\
  \texttt{danteev@amazon.com} \\
  \texttt{dante.everaert@gmail.com}
  \And
  Christopher Potts \\
  Stanford University \\
  \texttt{cgpotts@stanford.edu} \\
}

\begin{document}

\doparttoc 
\faketableofcontents 

\maketitle

\begin{abstract}
It is often advantageous to train models on a subset of the available train examples, because the examples are of variable quality or because one would like to train with fewer examples, without sacrificing performance. We present \ourmethod\ (\ourmethodabbrev), a scalable, task-agnostic approach to this data selection problem that requires only a small set of (unlabeled) examples representing a target distribution. \ourmethodabbrev\ begins from a natural, information-theoretic objective that is intractable in practice. Our contribution is in showing that it can be made highly scalable through a simple relaxation of the objective and a highly efficient implementation. In experiments with machine translation, spelling correction, and image recognition, we show that \ourmethodabbrev\ delivers outstanding results with very small train sets. These findings are robust to different representation models and hyperparameters for \ourmethodabbrev\ itself. \ourmethodabbrev\ is task- and domain-agnostic and can be applied out-of-the-box to new datasets and domains. We open source a pip-installable implementation of the algorithm as "pip install grad-info-opt".\footnote{\parbox [t] {\linewidth}{pip install grad-info-opt,  see \appref{app:algorithm} for details. Also available at: \\\url{https://github.com/daeveraert/gradient-information-optimization}\label{footnote_algo}} }
\end{abstract}

\section{Introduction}

In situations in which one has a very large train set available, it is often advantageous to train systems on a subset of the data. In the simplest case, the train set may be so large as to run up against resource constraints, and the question arises whether performance goals can be reached with less  effort \citep[e.g.][]{resource}. It can also be the case that the train examples are known to be of variable quality, say, because they were harvested from diverse websites \citep{cc}, annotated by crowdworkers \citep{crowd}, or created by a synthetic data generation process \citep{synthetic}. In this case, the goal is to identify a reliable subset of examples.


This is the data selection problem that we address in the current paper. The end goal is to select a subset of the available train examples  that leads to models that are at least as performant as (and perhaps even better than) those trained on all the examples. To achieve this goal, we propose \ourmethod\ (\ourmethodabbrev), a highly scalable, task-agnostic approach to data selection that is based in information theory. Our method assumes access to a (potentially small) set of examples~$X$ that represent the desired data distribution and a (presumably very large) set of potential train examples~$G$. Our method derives a set $V \subseteq G$ that has as much information content as possible about the target distribution $X$. The method begins from the natural intuition that we want $V$ to minimize the average KL divergence from $X$, and the novelty of the approach lies in making this computationally tractable by relying on properties of the derivative of the KL divergence and implementing the method extremely efficiently. Crucially, our method works in any continuous representation space, is task- and domain-agnostic, and requires no labels on examples.


We motivate \ourmethodabbrev\ with a diverse set of experiments. We first explore machine translation using the WMT14 dataset and Transformer-based models. In this case, $G$ is the WMT14 dataset and $X$ is the dev set. These experiments show that, using \ourmethodabbrev, we can can surpass the performance of a model trained on the full WMT14 corpus with only a fraction of the example in $G$, which represents very large efficiency gains. We then turn to spelling correction. In this case, the set $G$ is generated by a noisy synthetic process and the target distribution $X$ is a set of actual spelling errors. Here, we are using \ourmethodabbrev\ to home in on realistic train examples. Our results show that we can do this extremely effectively. Finally, we apply \ourmethodabbrev\ to an image recognition task (FashionMNIST) and show again that our method can reduce the size of the train sets chosen without large drops in performance, this time operating with representations of images. In this case, we trust the train set $G$ to represent the space accurately, and our goal is simply to select a useful subset of $G$. Thus, in this case $X = G$. Finally, we discuss expanding \ourmethodabbrev,  report on a wide range of robustness experiments and empirical analyses of how and why the method works in practice, and publish a pip-installable package. \footref{footnote_algo}
\section{Related Work}\label{gen_inst}

\paragraph{Active learning.}
Active learning methods \citep[e.g.][]{sener2018active, al2, al3} can be cast as data selection methods in our sense. In active learning, one iteratively chooses new unlabeled training examples to label, with the goal of efficiently creating a powerful train set. By contrast, \ourmethodabbrev\ makes no use of labels and is oriented towards the goal of identifying a subset of existing cases to use for training. Additionally, active learning is most suited to classification problems, whereas \ourmethodabbrev\ works with any arbitrary task.


\paragraph{Heuristic.} \ourmethodabbrev\ is closer to recent methods in which one uses a large language model to generate a large number of candidate texts and then extracts a subset of them based on a specific criteria. For example, 
\citet{brown} develop a heuristic method to filter CommonCrawl based on a trained classifier's probability that datapoints are high quality. Similarly, 
\citet{ccnet} develop a pipeline to clean CommonCrawl based principally on the perplexity of an LM trained on high quality text, and \citet{importance_resampling} develop a sampling technique based on approximate n-gram counts. 

Like \ourmethodabbrev, these heuristic methods aim to select a subset of data that is higher quality and more relevant. However, they are either highly tailored to their particular tasks or they require very large numbers of examples (to develop classifiers or construct target probabilities). By contrast, \ourmethodabbrev\ is task- and domain-agnostic, it can be applied plug-and-play to a new task and dataset, and it requires comparatively few gold examples $X$ to serve as the target distribution.


\paragraph{Similarity Search.} Methods using vector or n-gram similarity search can also be used for data selection at scale \citep[e.g.][]{faiss, annoy1, colbertv2}. The technique would index $G$ and $X$ and retrieve the top-k datapoints from $G$ for each point in $X$. Like our method, similarity search works in a continuous space. However, similarity search can be prone to selecting suboptimal points; we review such a case in detail in \secref{sec:benchmarks}. Additionally, similarity search does not have a natural stopping criterion and requires data size to be chosen arbitrarily. Is 10\% data enough? 20\%? We don't know a priori. And if the data in $G$ is far away from $X$, similarity search will still choose it up to the desired data size. Recently, \citet{yao22c} use a BM25 retrieval method for data selection, with strong results. However, BM25 operates on a bag-of-words model, which can make it challenging when the target set is small, and like any similarity search, requires data size to be chosen arbitrarily beforehand. Further, this method only applies to text tasks, whereas \ourmethodabbrev\ applies to any task with continuous representation.

\paragraph{Data Pruning.} Data pruning has shown promise  for the data selection problem \citep[e.g.][]{el2n}. Different works in data pruning use varied approaches, but generally iteratively identify and add optimal samples from training data. However, most works in data pruning \citep[e.g.][]{el2n,pruning_class,pruning_class2} apply only to classification. However, recently \citet{pruning} develop a self-pruning method with strong results that can be applied to any task with continuous representation, like \ourmethodabbrev, using clustering-based selection. Unlike \ourmethodabbrev\ and similarity search however, self-pruning does not consider any desired target  distribution $X$.

\paragraph{Submodular Optimization.} Submodular optimization methods have also been proposed for data selection \citep{submodlib}, where an optimizer optimizes a submodular function between a general set $G$ and a target set $X$. Like \ourmethodabbrev, submodular optimization takes into account a desired target distribution $X$, has a natural stopping criterion that does not require data size to be chosen arbitrarily. Unlike \ourmethodabbrev\ however, submodular optimization is restricted only to submodular functions, which have certain assumptions that may not hold true. For example, submodular functions assume adding each extra datapoint diminishes the return of that datapoint, which is not necessarily the case when constructing a dataset. For example, if we have already have selected data but have unexplored region where $X$ has a mode, adding data in that region does not exhibit a diminishing return. \ourmethodabbrev, on the other hand, makes no assumptions on the functions it can use.

Overall, previous work in data selection is typically tailored to a specific domain like NLP \citep[e.g.][]{ccnet} or image recognition \citep[e.g.][]{al2}, and makes assumptions about the data available, for example, that the target set $X$ is large enough to construct an LM \citep[e.g.][]{ccnet}, or that it has labels \citep[e.g.][]{sener2018active}. In addition, many of these methods use discrete approximations and heuristics \citep[e.g.][]{importance_resampling, heuristic-2}. In this work, we provide a general, theoretically-motivated data selection method that works with large or small $X$ and can be applied out-of-the-box to any domain (image, text, etc) without needing labels.

\input{algorithm}

\section{\ourmethod: Method}
\label{method}
We formulate data selection as maximizing information content and outline the natural algorithm for this objective, which is infeasible. We then introduce optimizations which enable the algorithm to work at scale, and conduct tests to show the algorithm is consistent and robust to different scenarios.


\subsection{Abstract Formulation of the Data Selection Problem} 

We assume that all examples are represented in continuous space. We have a set of train examples~$G$ and a target ideal state $X$. We allow also that there may be existing train examples $D$ that we definitely want to include in our train set, though $D$ can be empty. Our goal is to identify a subset $V$ of $G$ such that the set $D \cup V$ contains the most information about $X$.


In this setting, it is natural to take an information-theoretic approach. Let $p_X(\mathbf{x})$ be the distribution of target $X$, and let $p_{D \cup V}(\mathbf{x})$ be the distribution of selected data $D \cup V$. The information content of $D \cup V$ about $X$ is the negative KL divergence from $p_X(\mathbf{x})$ to $p_{D \cup V}(\mathbf{x})$ \citep{kldiv}. In this context, the general objective of data selection is as follows:
\begin{equation}\label{eq:kl}
\text{Choose data $V \subseteq G$ such that $\displaystyle \int_{\Omega}p_X(\mathbf{x}) \space \log \frac{p_X(\mathbf{x})}{p_{D \cup V}(\mathbf{x})} \space d\mathbf{x}$ is minimized}
\end{equation}
The implication is that a data selection method which gives the minimum KL divergence will also give the best performance (assuming we are correct that $X$ represents the task to be solved).

\subsection{Naive Approach}

A natural approach is to hill-climb on the KL divergence objective \eqref{eq:kl}. Given existing data $D$ and points $\mathbf{v}_1, \ldots, \mathbf{v}_k$ of $G$, we recompute the distribution $p_{D \cup \{\mathbf{v}_i\}}(\mathbf{x})$ for each  $\mathbf{v}_i$, pick the one that gives the minimum KL divergence, and add it to our selected set $D$:
\begin{equation}\label{eq:naive}
D \leftarrow D+\argmin_{\mathbf{v}_i \in G}\displaystyle \int_{\Omega} p_X(\mathbf{x}) \space \log \frac{p_X(\mathbf{x})}{p_{D \cup\{\mathbf{v}_{i}\}}(\mathbf{x})} \space  d\mathbf{x}
\end{equation}
Unfortunately, this algorithm is intractable in practice. We need to construct a new distribution $p_{D \cup \{\mathbf{v}_i\}}(\mathbf{x})$ and compute KL divergence for every $\mathbf{v}_i \in G$, at each step. Therefore, the complexity at each iteration is $\mathcal{O}( |G| \cdot C)$, where $C$ is the cost of computation for the KL divergence. For a dataset of only 1M and 0.1s per iteration, it would take 70 days to complete the algorithm. The method is also prone to adding the same point multiple times.


\subsection{\ourmethod}

\ourmethodabbrev\ addresses the shortcomings of \eqref{eq:naive} with a combination of mathematical and implementational optimizations. The method is described in Algorithm~\ref{alg:cap}.

First, instead of calculating divergence for each point, we use the derivative of the KL divergence to find the optimal point. We rewrite $p_{D \cup \{\mathbf{v}_i\}}(\mathbf{x})=g(\mathbf{x}, \mathbf{v}_i)$, a function of only $\mathbf{x}$ and $\mathbf{v}_i$ since $D$ is not changing, and thus the optimization term in each iteration becomes:
\begin{equation}\label{eq:original}
\argmin_{\mathbf{v}_i \in G} \int_{\Omega} p_X(\mathbf{x}) \space \log \frac{p_X(\mathbf{x})}{g(\mathbf{x}, \mathbf{v}_i)} \space  d\mathbf{x}
\end{equation}
We can relax the constraint that $\mathbf{v}_i \in G$ to the space of all possible $\mathbf{v}$ and solve this integral minimization for the optimal $\mathbf{v}_{\textit{opt}}$. Since $p_X$ is unchanging and the integral implicitly removes $\mathbf{x}$ as a variable, the integral defines a functional $F\left[g(\mathbf{v})\right]$ . Therefore, we partially differentiate with respect to $\mathbf{v}$ and do gradient descent with the partials $\nabla_{\mathbf{v}_k} F[g]$ to solve for $\mathbf{v}_{\textit{opt}}$. All together, this becomes:
\begin{equation}\label{eq:ours}
\mathbf{v}_{k+1} \leftarrow \mathbf{v}_k-\gamma \cdot \frac{\partial}{\partial \mathbf{v}_k}\left( \int_{\Omega} p(\mathbf{x}) \space \log \frac{p(\mathbf{x})}{g(\mathbf{x}, \mathbf{v}_k)} \space  d\mathbf{x}\right)
\end{equation}
Once we have $\mathbf{v}_{\textit{opt}},$ we find the nearest $\mathbf{v}_i \in G$ to add to $D$, as the closest $\mathbf{v}_i \in G$ is the solution to \eqref{eq:original}.
We assume $G$ is locally dense for the extrema of the integral in \eqref{eq:ours}; see \appref{app:local-density} for details. 

The complexity at each iteration, for $S$ gradient descent steps, is $\mathcal{O}(S \cdot C)$ which does not increase with $G$. Therefore, when $|G| > S$, as is common in practice, the derivative trick is faster than the naive algorithm. We time both algorithms in \secref{sec:benchmarks} and show the derivative trick is 80\% faster.

\input{explosion}

Second, even at its most efficient, an algorithm that adds point-by-point becomes intractable. Therefore, we use a quantization-explosion process. First, we cluster the data with K-means \citep{k-means} and pick the centroids $\boldsymbol{\mu}_i$. We then perform the algorithm using the cluster centroids $\boldsymbol{\mu}_i$ instead of the original data. Finally, we use our chosen cluster centroids and explode back  to the original data based on cluster membership. \Figref{fig:quantization-explosion} provides an overview of the process.

Third, to compute the KL Divergence in high-dimensional spaces, we use the k-Nearest Neighbors approximation for continuous KL divergence proposed by \citet{kldivest}, and modify it to be an average across all points to bypass 0 gradient problems (details and proof of modification are in the \appref{app:0-grad-kl-mod}). Let $|D|=m$, $|X|=n$ and $d$ be the dimensionality:
\begin{equation}
    \hat D_{\textit{KL}}(P_X \parallel P_D)=\frac{1}{m} \sum_{k=1}^{m}\frac{1}{n}\left[\sum_{i=1}^{n}d \cdot\log \nu_k(i)-d \cdot\log \rho_l(i)\right]+\frac{1}{m}\sum_{k=1}^{m} \log \frac{l\cdot m}{k(n-1)}
\end{equation}

Where $\nu_k(i)$ is the distance from point $X_i$ to the $k$th nearest point in $D$ and $\rho_l(i)$ is the distance from point $X_i$ to the $l$th nearest $X_{j \neq i}$. We use automatic differentiation to compute the derivative.

We can stop when the KL divergence increases, reset $G$ and allow the algorithm to pick again, among a variety of criteria. We explore several in our experiments and list additional criteria in \appref{app:algorithm:stop_crit}. Unlike data selection methods that make data size a hyperparameter \citep[e.g.][]{yao22c}, \ourmethodabbrev\ provides a natural stopping criterion (KL divergence). Finally, initializing $D$ from a uniform start rather than empty leads to same optimal points but a smoother convergence; see \appref{app:uniform-start}.

\paragraph{Limitations.} We derived \ourmethodabbrev\ from the natural information-theoretic objective, however, we can use any arbitrary statistical distance in the \ourmethodabbrev\ framework. For example, in situations where $G$ is close to $X$ with the exception of a large gap somewhere, the statistical distance $\max |p_X(\mathbf{x})-p_D(\mathbf{x})|$ may be better suited. We also use gradient descent to iteratively find $\mathbf{v}_{\textit{opt}}$, but we know the space is non-convex. Therefore, replacing gradient descent with a method like particle swarm optimization \citep{pso}  may lead to better selected data. Finally, in practice it is important to ensure that $X$ reasonably represents the space a model might be used on. A narrow $X$ could make a model trained on \ourmethodabbrev-selected data perform poorly when confronted with inputs that lie outside $X$. Methods like starting from a subset of training data, which we explore, or adding uniform points to $X$ to encourage generalization, should be explored. We leave these improvements to future work.

\subsection{Analytic Checks}\label{sec:benchmarks}
\input{consistency}
\paragraph{\ourmethodabbrev\ is self-consistent.} 
We define self-consistency as follows: if both $G$ and $X$ come from the same distribution, i.e., $p_G(\mathbf{x})=p_X(\mathbf{x})$, a good data selection method should choose all of $G$. We show \ourmethodabbrev\ is self-consistent: let $X$ be 100 points from a normal distribution centered at $(3,4)$ and let $G$ be another 100 points from the same distribution (\figref{fig:consistency}, 1\textsuperscript{st} graph). We run \ourmethodabbrev\ on this setup; \ourmethodabbrev\ selects 96\% of $G$ before termination, showing \ourmethodabbrev\ is self-consistent (\figref{fig:consistency}, 2\textsuperscript{nd} graph).

\paragraph{\ourmethodabbrev\ is negative-consistent.}  
We define negative consistency as follows: if $G$ is very far from $X$, i.e. $d(p_X(\mathbf{x}),p_G(\mathbf{x})) \gg 0$, a good data selection method should not choose any of $G$. Most data selection methods that rely on choosing a desired data size as a stopping criteria \citep[e.g.][similarity search]{yao22c, importance_resampling} are not negative consistent; they will select data regardless of how close or far the data may be from $X$. We show \ourmethodabbrev\ is negative-consistent with the following setup: let $X$ be the same as above, but this time let $G$ be 100 points centered far away at $(300, 400)$. We run \ourmethodabbrev\ on this setup; \ourmethodabbrev\ terminates without adding any points from $G$, showing it is negative-consistent.

\paragraph{Quantization in \ourmethodabbrev\ is consistent with the original space.} 
Quantizing the space with K-means should not change the distribution of data. We show this with the following setup: let $X$ be 400 points from a normal distribution centered at $(3,4)$. We quantize $X$ using K-means with K=50, and compute the KL divergence $\hat D_{\textit{KL}}(X \parallel X_{\textit{quant}})$, which should be near $0$ if the distributions are close. The KL divergence is $0.44$, showing the quantization in \ourmethodabbrev\ is consistent with the original space.

\paragraph{The derivative trick is 80\% faster.} We benchmark the wall-clock time between the naive hill-climb method and \ourmethodabbrev\ with the derivative trick, with 100 points in $X$ and 2000 points in $G$ spread uniformly, and run the algorithm for 100 iterations. The regular hill-climb method takes 1369s, whereas the derivative trick takes 257s, representing an 80\% speedup.

\paragraph{\ourmethodabbrev\ selects based on a distribution; similarity search does not.} Let $X$ be a circle of 2D points encapsulating a region of space, and let $G$ be 2000 uniformly distributed points (\Figref{fig:consistency}, 3\textsuperscript{rd} and 4\textsuperscript{th} graphs). The ideal points are the points within the region encapsulated by $X$ and should be chosen with preference over points outside. \Figref{fig:consistency} shows the points selected by semantic search (4\textsuperscript{th} graph); in order to get the data \textit{inside} the circle, it also picks the data \textit{outside} the circle. \Figref{fig:consistency} also shows the \ourmethodabbrev-selected points (3\textsuperscript{rd} graph); by considering the \textit{distribution} of all of $X$ rather than the simple Euclidean location of each $X$, \ourmethodabbrev\ selects mostly points which are within the circle, as desired.

\section{Experiments}

We perform four sets of experiments to validate \ourmethodabbrev. First, we replicate the setup of \citet{vaswani2017transformer} on WMT14 and show that using \ourmethodabbrev\ can surpass the performance of the full model with a fraction of the data. Next, we demonstrate \ourmethodabbrev\ is robust to different choices of embedding models and quantization. Third, we use a spelling correction task to show that \ourmethodabbrev\ selects the highest quality data from a pool of mixed-quality synthetic data. Finally, we show \ourmethodabbrev\ reduces the training set size of FashionMNIST image task without a big drop in performance. We show that \ourmethodabbrev\ achieves the lowest KL divergence compared to alternatives, and that this correlates with model performance.  \footnote{Details of each experiment are in \appref{app:experiments}}

\begin{table}[tp]
   \caption{Machine translation results. Training data sizes and BLEU scores of models trained on full data, \ourmethodabbrev-selected data, comparative methods and random subsets for various initialization states. \textbf{Bold} is the best score in each initialization, \textit{italic} is the best score overall. \ourmethodabbrev\ outperforms a model trained with the full EN-FR with only 40\% of the data and outperforms all comparative methods in 10/12 evaluations. It achieves 99\% of the performance in EN-DE with only 60\% of the data.}
  \label{tab:mt-results}
  \centering
  {\begin{tabularx}{\textwidth}{XXXXXXXXXX}
    \toprule
    \multirow{4}{*}{\textbf{Init. \%}} & \multirow{4}{*}{\textbf{System}} & \multicolumn{4}{c}{EN-FR} & \multicolumn{4}{c}{EN-DE} \\
    \cmidrule(lr){3-6} 
    \cmidrule(ll){7-10} 
    & & \textbf{Train Size} & \textbf{Dev Test} & \textbf{WMT 14} & $\mathbf{\hat D_{KL}}$ & \textbf{Train Size} & \textbf{Dev Test} & \textbf{WMT 14} & $\mathbf{\hat D_{KL}}$\\
    \midrule
     & Ours  & & \textbf{34.2} & \textbf{41.2} & 156 & & 22.1 & 24.3 & \textit{\textbf{148}}    \\
         & BM25 & & 33.9 & 41.0 & 172 &  & \textbf{22.6} & \textbf{24.9} & 175     \\
    0 & Pruning       & \phantom{0}5.6M & 33.3 & 40.3 & \textit{\textbf{152}} & 701K & 22.0 & 24.2 & 177  \\
    & Submod.       & & 33.8 & 40.8 & 170 & & 22.0 & 24.5 & 164  \\
         & Random       & & 33.1 & 40.0 & 194 & & 21.9 & 24.0 & 183  \\
    \midrule
         & Ours  & & \textit{\textbf{34.8}} & \textbf{42.2} & \textbf{166} & & \textbf{23.9} & \textbf{27.0} & \textbf{159}    \\
         & BM25 &  & 34.6 & 42.0 & 179 &  & 22.9 & 26.3 & 178     \\
    25    & Pruning  & \phantom{0}14M & 34.1 & 41.4 & 174 & 1.7M & 23.0 & 26.3 & 171  \\
    & Submod.       & & 34.4 & 41.7 & 181 & & 22.7 & 25.5 & 172  \\
         & Random       & & 34.3 & 41.4 & 195 & & 23.0 & 26.7 & 182  \\
    \midrule
             & Ours  & & \textbf{34.7} & \textit{\textbf{42.3}} & \textbf{172} & & \textit{\textbf{24.2}} & \textbf{27.9} & \textbf{164}    \\
         & BM25 &  & 34.3 & 42.1 & 185 &  & 23.7 & \textbf{27.9} & 178     \\
        50    & Pruning  & \phantom{0}21M & 34.2 & 41.9 & 183 & 2.5M & 23.6 & 27.1 & 179  \\
    & Submod.       & & 34.4 & 41.7 & 185 & & 21.9 & 24.5 & 176  \\
         & Random       & & 34.1 & 41.7 & 195 & & 24.0 & 27.3 & 181  \\
    \midrule
    100 & Full\footnotemark & 35M & - & 41.8 & 188 & 4M & - & \textit{28.2} & 180 \\
    \bottomrule
    
  \end{tabularx}}
\end{table}
\footnotetext{From \citet{vaswani2017transformer}}
\subsection{Machine Translation Experiments}\label{sec:wmt14}

Our first set of experiments seeks to show that \ourmethodabbrev\ can pick data from a general corpus to meet or exceed the performance of a model trained on the full corpus.


\paragraph{Data and Methods}

We use Transformer Big from \citet{vaswani2017transformer}, trained for 300k iterations with the same hyperparameters. We use the same processed WMT14 training data. We report the BLEU score \citep{papineni-bleu} on the WMT14 test set.

We apply \ourmethodabbrev\ to select a subset of data from the WMT14 train set using the inputs only (as our method makes no use of labels). $G$ is the training data we can select from. For target state $X$, we collect the dev sets for WMT08--WMT13, extract 3K pairs and report BLEU on this held out dev set, and use the remaining $\approx$12K pairs as $X$. For initial state $D$, we consider starting from an empty set, a 25\% random subset of train data, and a 50\% random subset of train data, and we report results for each setting. We use MPNet-Base-V2 model \citep{song2020mpnet} to embed the input sentences in a continuous vector space and use K=1500 for quantization. We compare this embedding model and quantization amount to other settings in our robustness experiments (\secref{sec:robustness}).  As our stopping criteria, we stop when the KL divergence increases. We also deduplicate the data pairs before training. 

We compare \ourmethodabbrev\ to a random subset of data of the same size. In addition, we compare against several competitive baselines with different approaches: \citeauthor{yao22c}'s \citep{yao22c} recent similarity search approach of BM25 retrieval, the recent data pruning self-pruning method introduced by \citet{pruning}, and the submodular information optimization with mutual information as the function to optimize. To keep the setup equal, we also initialize each of the baselines from a 0\%, 25\% and 50\% random subset and run those algorithms to have the same size as the \ourmethodabbrev-selected data. For submodular optimization and self-pruning, we use the same clustering \ourmethodabbrev\ is run with.

\paragraph{Results}

We find that \ourmethodabbrev\ outperforms the random baseline at every initialization. A data selection method should always outperform a randomly-selected subset of the same size. \Tabref{tab:mt-results}  shows the BLEU score on dev and WMT14 test sets\footnote{\label{FairseqNote}We use Fairseq's \citep{fairseq, fairseqgit} scripts} and demonstrates \ourmethodabbrev\ always outperforms random. BM25, submodular optimization and data pruning only outperform random sometimes.

\ourmethodabbrev\ outperforms the EN-FR model trained on the full data using only 40\% of the data. At initialization of 25\% and 50\%, a model trained on \ourmethodabbrev-selected data outperforms the full \citet{vaswani2017transformer} model trained on all data by +0.4 and +0.6 BLEU, respectively. In addition, a model trained on the \ourmethodabbrev-selected data at 0\% initialization achieves 99\% of the performance of the full model with only 16\% of the data. It outperforms all comparative methods at all initializations. In EN-DE, it gets 99\% of the performance with 60\% of the data, and 88\% of the performance with only 18\% of the data.

\begin{table}[tp]
  \caption{Training data sizes and BLEU scores of the base version (K=1500 and MPNet) and the variants of K and embedding model. BLEU scores of the variants vary by only 0.4\% on average from the base, indicating \ourmethodabbrev\ is robust to different quantization and embedding models}
  \label{tab:robustness-results}
  \centering
  {\begin{tabularx}{\textwidth}{XXXXXXXXXX}
    \toprule
    \multicolumn{2}{c}{\multirow{4}{*}{\textbf{System}}} & \multicolumn{4}{c}{EN-FR} & \multicolumn{4}{c}{EN-DE} \\
    \cmidrule(lr){3-6} 
    \cmidrule(ll){7-10}
    & & \textbf{Train Size} & \textbf{Dev Test} & \textbf{WMT 14} & $\mathbf{\hat D_{KL}}$ & \textbf{Train Size} & \textbf{Dev Test} & \textbf{WMT 14} & $\mathbf{\hat D_{KL}}$\\
    \midrule
     \multicolumn{2}{l}{Base (MPNet, K=1500)} & 5.6M & 34.2 & 41.2 & 156 & 701K & 22.1 & 24.3 & 148    \\
    \midrule
    \multicolumn{2}{l}{\hskip0.75em MiniLM Variant} & 5.7M & 34.0 & 41.1 & - & 737K & 22.3 & 24.6 & -    \\
    \multicolumn{2}{l}{\hskip0.75em K=1000 Variant} & 5.6M & 33.9 & 41.2 & 169 & 701K & 22.1 & 24.3 & 150     \\
    \multicolumn{2}{l}{\hskip0.75em K=3000 Variant}      & 5.7M & 34.2 & 41.3 & 138 & 718K & 22.3 & 24.6 & 133  \\
    \midrule
    \multicolumn{2}{l}{Average Variance from Base} & 1.9\% & 0.5\% & 0.2\% & - & 3.7\% & 0.6\% & 0.8\% & - \\
    \bottomrule
  \end{tabularx}}
\end{table}

\ourmethodabbrev\ outperforms all comparative methods in 10/12 of the evaluations. \ourmethodabbrev\ always matches or outperforms the comparative methods with initializations of 25\% or 50\%, by an average of +0.7 BLEU on WMT14 and +0.6 BLEU on Dev Test, and only falls short
with 0\% initialization in EN-DE.

\ourmethodabbrev\ has the lowest KL divergence in 5/6 tests (\tabref{tab:mt-results}), which correlates with  model performance. The implication of \eqref{eq:kl} is that a dataset with lower KL divergence between train and target will perform the best. From \tabref{tab:mt-results}, the average Spearman rank correlation coefficient between KL divergence and best performance is 0.83 and the median is 1, showing a high degree of correlation between a dataset that minimizes KL divergence and model performance, and thus confirming the implication.

In summary, \ourmethodabbrev\ leads to the lowest KL divergence between train and target set out of all the methods, which correlates with model performance and confirms the theory in \eqref{eq:kl}. Notably, a model trained with \ourmethodabbrev-selected data outperforms a model trained on the full data in EN-FR despite using only 40\% of the total data and came to within 99\% of the full model in EN-DE using only 60\% of the data.  \ourmethodabbrev\ outperforms the random baseline at all initializations and outperforms all comparative methods in 10/12 evaluations. Overall, these experiments show \ourmethodabbrev\ can achieve and surpass the performance of a model trained on full data and comparable baselines, by explicitly optimizing for KL Divergence.
\subsection{Robustness}\label{sec:robustness}

\ourmethodabbrev\ above relies on two approximations to work: an embedding model, and K-means to quantize the space into representative samples of the full data. In this section, we show that \ourmethodabbrev\ is robust to different embedding models and different values of K. The results are summarized in \Tabref{tab:robustness-results}.

\paragraph{\ourmethodabbrev\ works with different embedding models.} \ourmethodabbrev\ should be robust to different text embedding models. We change the embedding model from MPNet-Base-V2 to MiniLM-L12-v1 \citep{wang2020minilm}, which has different architecture, training, and produces embeddings of a different size. We then rerun the 0\% initialization experiments end-to-end with the new embeddings for both EN-DE and EN-FR. \Tabref{tab:robustness-results} shows that using MiniLM in \ourmethodabbrev\ results in roughly similar selected data size (4.4\% difference on average) and virtually identical performance (0.7\% difference on average), demonstrating that \ourmethodabbrev\ is robust to different embedding models.

\begin{table}[tp]
 \caption{Spelling correction results. Training data size, \% high quality data and KL Divergence of the full data, \ourmethodabbrev, BM25, submodular optimization, data pruning and random data. \textbf{Bold} is the best score overall. \ourmethodabbrev\ selects 73\% high quality data which outperforms all other methods.}
  \label{table-speller}
  \centering
  {\begin{tabular}{l c c c}
    \toprule
    \textbf{System} & \textbf{Train Size} &  \textbf{\%High Quality} & $\mathbf{\hat D_{KL}}$ \\
    \midrule
    Ours  & &  \textbf{73\%} & \textbf{224}    \\
    BM25 &  & 55\% & 264    \\
    Pruning & \phantom{0}3.6M  & 54\% & 241    \\
    Submod. &  & 59\% & 245    \\
    Random  & &  50\% & 284  \\
    \midrule
    Full & 14.7M &  50\% & 280 \\
    \bottomrule
  \end{tabular}}
\end{table}

\paragraph{\ourmethodabbrev\ works with different choices of K.}  \ourmethodabbrev\ should also be robust to varying amounts of quantization. We decrease the value of K from 1500 to 1000 and increase to 3000 and rerun the 0\% initialization experiments end-to-end for both new values of K, in EN-FR and EN-DE. For K=1000 due to the coarser grain, \ourmethodabbrev\ selects more data, therefore we sample from the selected data the same amount as K=1500 in order to maintain parity. \Tabref{tab:robustness-results} shows that performance is virtually identical between the different values of K (0.4\% difference on average), demonstrating \ourmethodabbrev\ is robust to different values of K. In general, higher values of K have lower KL Divergence and slightly better performance, which is expected as the quantization is more fine grained.


\subsection{Spelling Correction Experiments}\label{sec:speller}

In this section, we set up a problem with a pool of high and low quality synthetic candidate train examples and show \ourmethodabbrev\ selects mostly high quality data. In addition, we set a new state of the art on the challenging BEA4660 spelling correction benchmark of \citet{neuspell}; see \appref{app:experiments}.

\paragraph{Data and Methods}

We follow the setup of \citet{neuspell} and collect 15M samples from the 1 Billion Word benchmark corpus and deduplicate. To create high quality data, we use the best noising technique (prob) from \citet{neuspell} and noise half the data. For low quality data, we use the “word” method with high replacement rate (70\%) and noise the other half, and mix the two sets.


We apply \ourmethodabbrev\ to select a subset of data from the training set. $G$ is the training data we can select from. For target state $X$, \citet{neuspell} provide 40k real spelling mistakes and corrections from the BEA grammar correction corpus. For initial state $D$, we start from an empty set. For the embedding model, we use MPNet and K=1500 for quantization. As our stopping criteria, we experiment with a new scheme: first the algorithm runs until the KL divergence increases, then we reset $G$ and allow the algorithm to pick again from the training data, until the KL divergence decreases. As before, we compare \ourmethodabbrev\ to a random subset of the same size and pruning, submodular and BM25 methods.


\paragraph{Results}

\ourmethodabbrev\ selects high quality data. A good data selection method should select mostly from the high quality data. \tabref{table-speller} shows \ourmethodabbrev\ selects 73\% high quality data, compared to 59\% for submodular optimization, 55\% for BM25 and 54\% for pruning. \ourmethodabbrev's KL Divergence is lower than comparative methods and random, indicating KL divergence is also an indicator of data quality in this setup.


\subsection{Image Recognition}\label{sec:fashionmnist}

\begin{table}[tp]
 \caption{Image recognition results. Training data sizes and accuracy of models trained on \ourmethodabbrev\-selected data and random subset. \textbf{Bold} is the best score between ours and random. \ourmethodabbrev\ gives the best performance under the reduction in training data size. Full model is provided for comparison}
 \label{tableimg}
  \centering
  \setlength{\tabcolsep}{12pt}
  {\begin{tabular}{l c c r}
    \toprule
    \textbf{System} & \textbf{Size Train/Valid} & \textbf{Accuracy\footnotemark} & $\mathbf{\hat D_{KL}}$\\
    \midrule
    Ours  & 15,000/1,700 & \textbf{92.2\%} & 771\hspace{4pt} \\
    Random  & 15,000/1,700 & 91.5\% & 740\hspace{4pt} \\
    \midrule
    Full & 56,300/3,700 & 94.5\% & 739\footnotemark \\
    \bottomrule
  \end{tabular}}

\end{table}
\footnotetext[5]{Results averaged over 2 runs}
\footnotetext{See \appref{app:0-grad-kl-mod} for why this is not 0}
For our fourth set of experiments, we seek to show that \ourmethodabbrev\ works well in domains outside of NLP. We focus on the FashionMNIST \citep{fashionmnist} image recognition problem and show that we can use \ourmethodabbrev\ to dramatically reduce train set sizes without big drops in performance.


\paragraph{Data and Methods}

The FashionMNIST task has 10 classes of 28x28x1 images. There are 60,000 images in the training set, and 10,000 images in the test set. Our task will be to select a subset no more than 25\% of the total data that best approximates the training data. We then finetune the Resnet50 model \citep{resnet} for 5 epochs with Adam \citep{adam} (LR=5e-5) with the chosen data to do FashionMNIST classification. We split the data into train and validation sets, and pick the best checkpoint by validation loss. We report the accuracy on the test set.

We apply \ourmethodabbrev\ to select a subset of data from the training set. We use the training set as both $G$ and also as our target set $X$. We start $D$ from an empty set. We use the normalized and normed vector format of the images themselves in the algorithm, and use K=1000 for quantization. As our stopping criteria, we run the algorithm until we get 250 clusters (250 iterations), which is \textasciitilde25\% of the data. We also report results on a random sample of 25\% of the data as a comparison.

\paragraph{Results}

\ourmethodabbrev\ outperforms a simple random subset by +0.7\%. \ourmethodabbrev\ in this setup is optimized to pick the images which add the most information about the entire training set. \tabref{tableimg} shows training on \ourmethodabbrev-selected data only dropped performance by 2.3\% from the full model, compared to a drop of 3.0\% for a random subset of the same size.


\section{Conclusion}
We presented \ourmethodabbrev, a task- and domain-agnostic data selection method that works in any continuous space with few assumptions. \ourmethodabbrev\ begins with the natural objective of minimizing KL divergence between a target set and selected set, and uses the gradient of KL divergence and an efficient implementation to find and select the data points that optimize that objective at scale. \ourmethodabbrev\ selected high quality data, selected data that outperformed models trained on full data and on recent data selection techniques, and was able to effectively reduce the training set size under a given resource constraint without big drops in performance. Current models consume large quantities of data, and we hope \ourmethodabbrev\ can help improve the performance of models while using less data and fewer resources. Additionally, with large quantities of synthetic and scraped data of variable quality available, we hope \ourmethodabbrev\ can help home in on high quality data. For example, \ourmethodabbrev\ can be used to select high quality synthetic data output by large language models for a particular task, or choose training data for LLMs that aligns with a certain intent. Improvements and changes to the statistics and optimization in \ourmethodabbrev\ and applications of \ourmethodabbrev\ to varied domains and tasks are promising directions for future work.



\section*{Acknowledgements}
We would like to acknowledge and thank the broader Amazon Search Spelling Correction and Autocomplete teams for their support during this work. We would like to thank in particular Parivesh Priye, Jingfeng Zhang, Xuan Guo, Alex Klementiev and Rohit Patki for their help in reviewing and giving invaluable feedback and insights over the course of developing the idea. We are grateful to all those who supported us and gave us feedback throughout, and apologize in advance if we missed anyone in this section.

\section*{Reproducibility} 
We include all the necessary code to reproduce both our method as well as all the baselines in the supplementary materials under the package gradient-information-optimization. Further, we include detailed instructions on the usage of all the code, including comparative method setup, as well as details on all hyperparameters used and testing setup in \appref{app:experiments}.

\bibliographystyle{plainnat}
\bibliography{iclr2024_conference}
\clearpage

\appendix
\addcontentsline{toc}{section}{Appendix} 
\part{Appendix} 
\parttoc 
\clearpage
\section{Theoretical Work}\label{app:theoretical}

\subsection{0 Gradient Problem and KL Divergence Modification}\label{app:0-grad-kl-mod}
\subsubsection{KL Divergence Estimator: Recap}
\citet{kldivest} propose a KL divergence estimator based on k-nearest neighbors (kNN) of points drawn from the probability density functions. We recap their derivation to provide necessary context for the eventual modification: Let $X=\{X_1 ... X_n\}$ be samples drawn from distribution $p$ and $Y=\{Y_1 ... Y_m\}$ be samples drawn from distribution $q$. Then the kNN estimate of $p$ at $X_i$ is:  
\begin{equation}\label{eq:p}
    \hat p_k(\mathbf{X}_i)= \frac{k}{n-1} \cdot \frac{1}{c_1(d) \rho_k^d(i)}
\end{equation}
Where $\rho_k(i)$ is the distance from $X_i$ to the $k$-nearest $X_{j \neq i}$, and $c_1(d)$ is the volume of the unit ball in the $d$-dimensional space. Likewise, the kNN estimate of $q$ at $X_i$ is:  
\begin{equation}\label{eq:q}
    \hat q_k(\mathbf{X}_i)= \frac{k}{m} \cdot \frac{1}{c_1(d) \nu_k^d(i)}
\end{equation}
Where $\nu_k(i)$ is the distance from $X_i$ to the $k$-nearest $Y_i$, and $c_1(d)$ is the volume of the unit ball in the $d$ dimensional space. They also propose, by the law of large numbers, that the estimate of KL divergence $\hat D_{\textit{KL}}(p \parallel q)$ is:
\begin{equation}\label{eq:large-numbers}
    \hat D_{\textit{KL}}(p \parallel q)=\frac{1}{n}\sum_{i=1}^n \log \frac{\hat p_k(\mathbf{X}_i)}{\hat q_k(\mathbf{X}_i)}
\end{equation}
Putting \eqref{eq:p}, \eqref{eq:q} and \eqref{eq:large-numbers} together, we get the \citet{kldivest} kNN estimator for KL divergence:
\begin{equation}\label{eq:kl-div-est}
\begin{aligned}
    \hat D_{\textit{KL}}(p \parallel q) {} & =\frac{1}{n}\sum_{i=1}^n \log \frac{\frac{k}{n-1} \cdot \frac{1}{c_1(d) \rho_k^d(i)}}{\frac{k}{m} \cdot \frac{1}{c_1(d) \nu_k^d(i)}}=\frac{1}{n}\sum_{i=1}^n \log \frac{m \cdot \nu_k^d(i)}{(n-1) \cdot \rho_k^d(i)}\\
    & =\frac{1}{n}\sum_{i=1}^n \left[d \log \nu_k(i) - d \log \rho_k(i)\right] + \log \frac{m}{n-1}
\end{aligned}
\end{equation}
\subsubsection{0 Gradient Problem}
However, \ourmethodabbrev\ uses the gradient of the estimator to find $\mathbf{v}_{opt}$ by computing $\frac{\partial}{\partial \mathbf{v}}\hat D_{\textit{KL}}\left(p_X \parallel p_{D \cup \{\mathbf{v}\}}\right)$, in other words, finding how the value of the estimator changes with a change in $\mathbf{v}$. We then run into the 0 Gradient Problem with the following scenario. Suppose $\mathbf{v}$ is far from all points in $X$. Then, adding $\mathbf{v}$ does not change the value of $\nu_k(i)$ (the distance from $\textbf{X}_i$ to the $k$th nearest point in $D$) as the closest points in $D$ to each point in $X$ are the same before and after adding $\mathbf{v}$. Therefore, the only term that changes in the KL divergence estimator is $m$, a constant, and when computing the derivative, the constant goes to 0. Altogether, this means that for points $\mathbf{v}$ that are far from $X$ and their $k$th nearest neighbors in $D$, the gradient will be 0 and we cannot do gradient descent. Formally:

Suppose we have $X$, $D$ and a new point $\mathbf{v}$, and are estimating $\hat D_{\textit{KL}}\left(p_X \parallel p_{D \cup \{\mathbf{v}\}}\right)$. In $\hat D_{\textit{KL}}$, $\nu_k(i)$ is the distance from $\mathbf{X}_i$ to the $k$th nearest neighbor in $D$, and $\rho_k(i)$ is the distance from $X_i$ to the $k$-nearest $X_{j \neq i}$. Suppose that $\forall \mathbf{x} \in X, \forall \mathbf{y} \in D, \parallel \mathbf{x} - \mathbf{v} \parallel\ >\ \parallel \mathbf{x} - \mathbf{y}_k \parallel+ \ \epsilon$ for a small $\epsilon$, where $\mathbf{y}_k$ is the $k$th nearest $\mathbf{y}$ in $D$ for a $k<K\in \{1...m\}$. Then, the partial gradient of $\hat D_{\textit{KL}}$ with respect to $\mathbf{v}$ as calculated in \ourmethodabbrev\ is:
\begin{equation}
    \frac{\partial}{\partial \mathbf{v}}\hat D_{\textit{KL}}\left(p_X \parallel p_{D \cup \{\mathbf{v}\}}\right)=\frac{d}{n}\sum_{i=1}^n \frac{\partial}{\partial \mathbf{v}} \log \nu_k(i)
\end{equation}
As no other terms depend on $\mathbf{v}$. Next, let $f(\mathbf{v})=\log \nu_k(i)$. We are then fundamentally computing $\displaystyle \lim_{\epsilon \rightarrow 0} \frac{f(\mathbf{v}+\epsilon)-f(\mathbf{v})}{\epsilon}$.
However, because $\forall \mathbf{x} \in X, \forall \mathbf{y} \in D, \parallel \mathbf{x} - \mathbf{v} \parallel\ >\ \parallel \mathbf{x} - \mathbf{y}_k \parallel+ \ \epsilon$, then $\nu_k(i)$ with $\mathbf{v}$ and $\nu_k(i)$ with $\mathbf{v} + \epsilon$ are the same, and therefore $f(\mathbf{v}+\epsilon)=f(\mathbf{v})$. But then:
\begin{equation}
    \lim_{\epsilon \rightarrow \mathbf{0}} \frac{f(\mathbf{v}+\epsilon)-f(\mathbf{v})}{\epsilon}=\lim_{\epsilon \rightarrow \mathbf{0}} \frac{\mathbf{0}}{\epsilon}=\mathbf{0}
\end{equation}
And we have the 0 Gradient Problem. Note: division and addition symbols above are taken element-wise in the vector space.

\subsubsection{KL Divergence Modification}
We now modify the KL divergence estimator to overcome the 0 gradient problem. In order to ensure that $\nu_k(i)$ with $\mathbf{v}$ and $\nu_k(i)$ with $\mathbf{v} + \epsilon$ are never the same, we take the average of estimators from $k=1$ to $k=m$ ($m$ is the size of $D$), i.e. calculate the divergence estimator from $X$ to \textit{every} point in $D$. In fact, \citet{kldivest} also recommend taking an average across different values of $k$, as a method to reduce variance. We formally derive this modification. Let $\nu_k(i)$ be the distance from $\mathbf{X}_i$ to the $k$th nearest neighbor in $D$, and $\rho_l(i)$ be the distance from $X_i$ to the $l$-nearest $X_{j \neq i}$. The mean of the estimator in \eqref{eq:kl-div-est} across all values of $k$ ranging from $1$ to $m$ is:
\begin{equation}\label{eq:kl-div-est-act}
\begin{aligned}
    \hat D_{\textit{KL}}(p \parallel q) {} &  =\frac{1}{m}\sum_{k=1}^m \left( \frac{1}{n}\sum_{i=1}^n \left[d \log \nu_k(i) - d \log \rho_l(i)\right] + \log \frac{l \cdot m}{k(n-1)}\right)\\
    & = \frac{1}{m}\sum_{k=1}^m \frac{1}{n}\sum_{i=1}^n \left[d \log \nu_k(i) - d \log \rho_l(i)\right] + \frac{1}{m}\sum_{k=1}^m\log \frac{l \cdot m}{k(n-1)}
\end{aligned}
\end{equation}
Given this modified estimator, we show that we no longer have the 0 gradient problem. Let us have the same setup, i.e.  $\forall \mathbf{x} \in X, \forall \mathbf{y} \in D, \parallel \mathbf{x} - \mathbf{v} \parallel\ >\ \parallel \mathbf{x} - \mathbf{y}_k \parallel+ \ \epsilon$, and further suppose this is true for all $k=1 ... m$. Thus, the derivative is:
\begin{equation}
    \frac{\partial}{\partial \mathbf{v}}\hat D_{\textit{KL}}\left(p_X \parallel p_{D \cup \{\mathbf{v}\}}\right)=\frac{1}{m+1} \sum_{k=1}^{m+1} \frac{d}{n}\sum_{i=1}^n \frac{\partial}{\partial \mathbf{v}} \log \nu_k(i)
\end{equation}
From before, we have shown that for all $k=1...m$, we arrive at a 0 gradient. However, this expression also has the term $\frac{\partial}{\partial \mathbf{v}} \log \nu_{m+1}(i)$, which mandatorily describes the distance from $\mathbf{X}_i$ to $\mathbf{v}$, as this value describes the distance from $\mathbf{X}_i$ to the \textit{furthest} point in $D\cup\{\mathbf{v}\}$. Therefore, we also have that $\parallel \mathbf{X}_i - (\mathbf{v}+\epsilon) \parallel\ \neq\ \parallel \mathbf{X}_i - \mathbf{v} \parallel$ and therefore $\nu_{m+1}(i)$ with $\mathbf{v}$ cannot equal $\nu_{m+1}(i)$ with $\mathbf{v}+\epsilon$. As a result, $f(\mathbf{v}+\epsilon)\neq f(\mathbf{v})$ and $\displaystyle \lim_{\epsilon \rightarrow \mathbf{0}} \frac{f(\mathbf{v}+\epsilon)-f(\mathbf{v})}{\epsilon} \neq \mathbf{0}$, as desired, and we avoid the 0 gradient problem. This is the estimator that \ourmethodabbrev\ uses in practice. 

We note that with the modified estimator, $\hat D_{\textit{KL}}\left(p_X \parallel p_X \right)\neq 0 $, as the first probability estimator is not averaged over $k$ but the second probability estimator is. In general, this is not an issue, as we are concerned with minimizing the KL divergence, but not concerned with the magnitude of the values themselves.

\subsection{Local Density Assumption}\label{app:local-density}
The derivative trick in \ourmethodabbrev\ finds a $\mathbf{v}_{opt}$ and then finds the closest $\mathbf{v}_i \in G$ to add to the selected data, as the closest $\mathbf{v}_i \in G$ is the solution to the original objective \eqref{eq:original}:
\begin{equation}\label{eq:original2}
\argmin_{\mathbf{v}_i \in G} \int_{\Omega} p_X(\mathbf{x}) \space \log \frac{p_X(\mathbf{x})}{g(\mathbf{x}, \mathbf{v}_i)} \space  d\mathbf{x}
\end{equation}
This holds true under the assumption of local density of the extrema of that integral, or more specifically, the minima of that integral. We outline the assumption:
\begin{equation}
\begin{aligned}
    {} & \text{\textit{Local Density Assumption}: Let $\mathbf{v_{opt}}$ represent the global minimum of $\displaystyle \int_{\Omega} p_X(\mathbf{x}) \space \log \frac{p_X(\mathbf{x})}{g(\mathbf{x}, \mathbf{v}_i)} \space  d\mathbf{x}$.}\\ 
    & \text{Then, for the closest $\mathbf{v}_i \in G$ to $\mathbf{v}_{opt}$, $\mathbf{v}_{b}$, to also be the solution to the original \eqref{eq:original2}, we assume}\\ 
    & \text{$\exists \mathbf{v}_b \in G\ \text{s.t.}\ \displaystyle \int_{\Omega} p_X(\mathbf{x}) \space \log \frac{p_X(\mathbf{x})}{g(\mathbf{x}, \mathbf{v}_b)} \space  d\mathbf{x} \leq \displaystyle \int_{\Omega} p_X(\mathbf{x}) \space \log \frac{p_X(\mathbf{x})}{g(\mathbf{x}, \mathbf{v}_j)} \space  d\mathbf{x}\ \forall \mathbf{v}_{j \neq b} \in G $.}
\end{aligned}
\end{equation}

In essence, for the closest $\mathbf{v}_i \in G$ to $\mathbf{v}_{opt}$ to also be the best solution, there needs to exist a point in $G$ close enough to the global minimum such that that point represents the greatest decrease in divergence if it were added over all other points. In practice, this is almost always true; if $G$ covers the space of $X$ well and has many points in it, which is true if $G$ is large enough, then this assumption is likely satisfied. We note that additionally, we assumed $\mathbf{v}_{opt}$ to represent the global minimum, but as we outlined in \secref{method} limitations, since we use gradient descent on a non-convex space this is not guaranteed to be true. We leave improvements of finding a more global $\mathbf{v}_{opt}$ to future work, as described in \secref{method} limitations. Even with gradient descent, it may be possible to attain more global values by simultaneously gradient-descending different parts of the space and picking the final $\mathbf{v}_i \in G$ based on lowest KL divergence.

Note that even if the local density assumption is violated, the results are not necessarily catastrophic. Even if we add a point that is the second or third best (or $n$th best, up to a limit), we are likely to continue to attain very close KL divergence values to if we followed the ideal optimization trajectory defined by \eqref{eq:original2}.
\begin{figure}[b]
    \centering
    \subfloat[With uniform start]{{\includegraphics[width=0.45\textwidth]{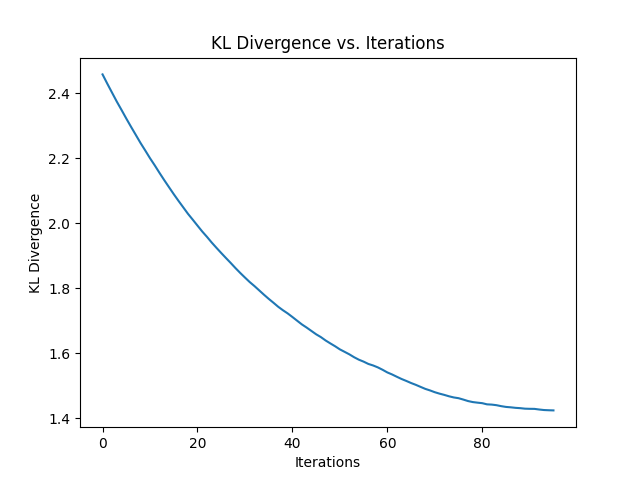} }}
    \subfloat[No uniform start]{{\includegraphics[width=0.405\textwidth]{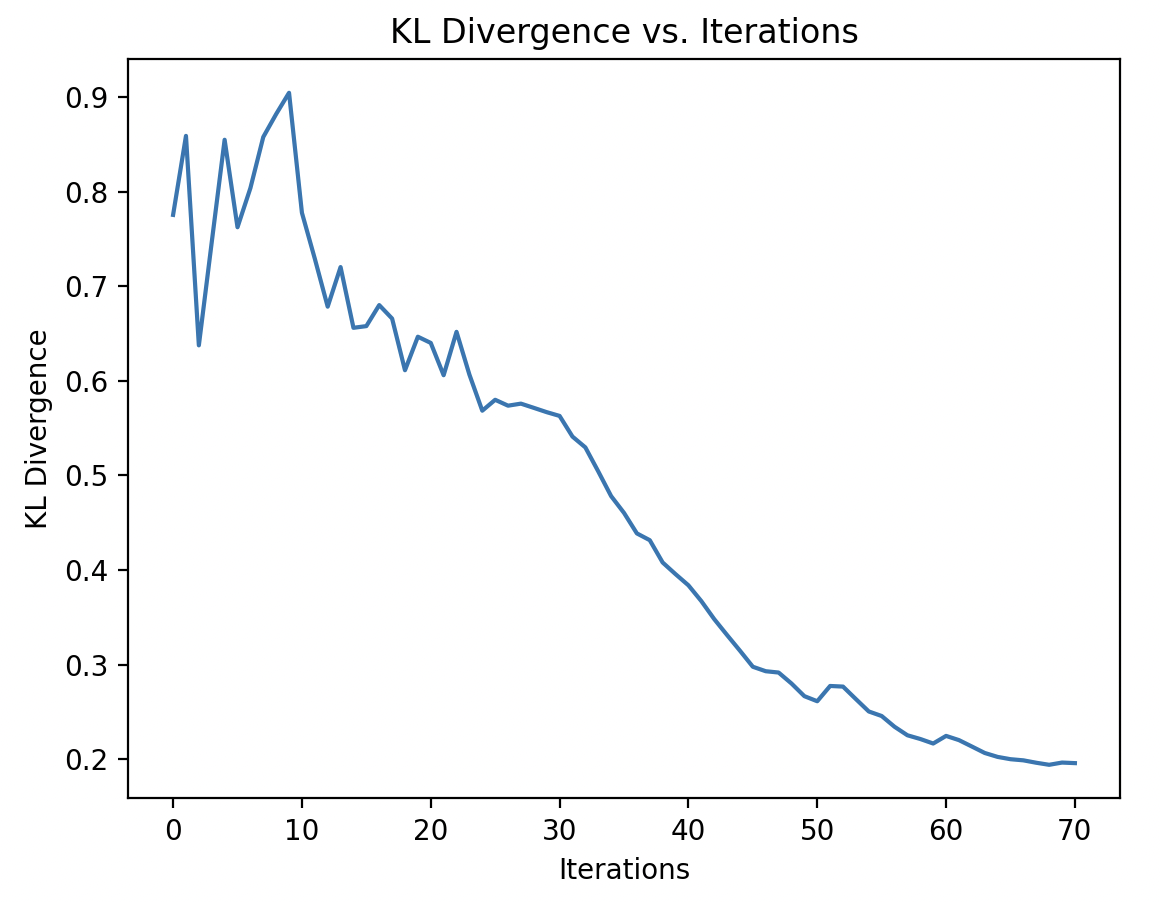} }}
    \caption{Left: KL divergence over iterations with uniform start, Right: KL divergence over iterations without uniform start. They follow the same trajectory (except the erratic behavior at the beginning), but adding a uniform start greatly improves the stability of the algorithm and convergence.}%
    \label{fig:uniform_comp}
\end{figure}
\subsection{Uniform Start}\label{app:uniform-start}
In order to smooth the convergence of the algorithm when starting from an empty set $D=\{\}$, we propose beginning with a uniform random spread of points as a good initialization basis. Intuitively, the uniform prior is the completely uninformative prior \citep{etjaynes}, and therefore does not change the values of $\mathbf{v}_{opt}$. We prove this property formally. Let $U(\mathbf{x})=A$ be a uniform distribution with fixed, $\mathbf{x}\text{-independent}$ probability $A$. Then, let us initialize $D \sim U(\mathbf{x})$, and then the resulting KL divergence optimization objective is a mixture of the uniform prior and whichever distribution of new points the algorithm is adding: $\frac{1}{c_1} A + \frac{1}{c_2} g(\mathbf{x}, \mathbf{v})$. In the regular version \eqref{eq:ours}, the gradient to find $\mathbf{v}_{opt}$ is:
\begin{equation}\label{eq:before}
\begin{aligned}
    {} & \frac{\partial}{\partial \mathbf{v}} \int_{\Omega} p_X(\mathbf{x}) \space \log \frac{p_X(\mathbf{x})}{g(\mathbf{x}, \mathbf{v})} \space  d\mathbf{x} \equiv \int_{\Omega} p_X(\mathbf{x}) \space \frac{\partial}{\partial \mathbf{v}} \log g(\mathbf{x}, \mathbf{v}) \space  d\mathbf{x}=\int_{\Omega} p_X(\mathbf{x}) \space \frac{\frac{\partial}{\partial \mathbf{v}} g(\mathbf{x}, \mathbf{v})}{g(\mathbf{x}, \mathbf{v})} \space  d\mathbf{x}
\end{aligned}
\end{equation}
With the uniform start expressed as a mixture, the gradient to find $\mathbf{v}_{opt}$ is:
\begin{equation}\label{eq:after}
\begin{aligned}
    \frac{\partial}{\partial \mathbf{v}} \int_{\Omega} p_X(\mathbf{x}) \space \log \frac{p_X(\mathbf{x})}{\frac{1}{c_1} A + \frac{1}{c_2} g(\mathbf{x}, \mathbf{v})} \space  d\mathbf{x} {} & \equiv   \int_{\Omega} p_X(\mathbf{x}) \space \frac{\partial}{\partial \mathbf{v}} \log \left(\frac{1}{c_1} A + \frac{1}{c_2} g(\mathbf{x}, \mathbf{v})\right) \space  d\mathbf{x}\\
    & =\int_{\Omega} p_X(\mathbf{x}) \space \frac{\frac{\partial}{\partial \mathbf{v}} g(\mathbf{x}, \mathbf{v})}{B + g(\mathbf{x}, \mathbf{v})} \space  d\mathbf{x}
\end{aligned}
\end{equation}
Where we divide by the constant $\frac{1}{c_2}$ on numerator and denominator and set $B=\frac{1}{c_1}A\cdot c_2$, another constant. The only difference between the gradient in \eqref{eq:before} and \eqref{eq:after} is that the denominator $g(\mathbf{x}, \mathbf{v})$ has a constant added to it. But then, it is possible to create an $\mathbf{x}$- and $\mathbf{v}$-independent bijection from \eqref{eq:before} to \eqref{eq:after} via $f:g(\mathbf{x}, \mathbf{v}) \rightarrow g(\mathbf{x}, \mathbf{v}) + B$. Since we are optimizing over $\mathbf{v}$ and implicitly $\mathbf{x}$ defined by $\Omega$, neither of which appear in the mapping, then the value $\mathbf{v}_{opt}$ that optimizes \eqref{eq:before} also optimizes \eqref{eq:after}, and therefore adding a uniform start does not change the value of $\mathbf{v}_{opt}$, as desired. Intuitively, the uniform prior adds a constant $B$ to $g$ in the derivative, and adding a constant to a function does not change the location of the extrema or the shape of that function. We give an example (from self-consistency experiment):

\section{Algorithm}\label{app:algorithm}
\begin{figure}[tp]%
    \centering
    \subfloat{{\includegraphics[width=0.45\textwidth]{Figure_1.png} }}
    \subfloat{{\includegraphics[width=0.45\textwidth]{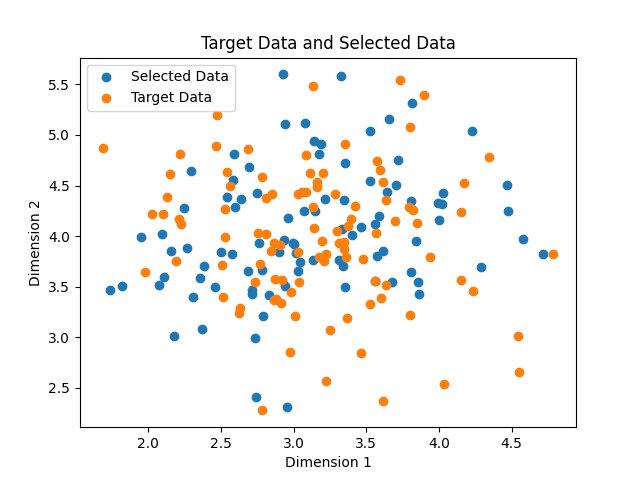} }}
    \caption{Sample plots for 2D example (Self-Consistency). Left: KL divergence over iterations, Right: Chosen data and target data}%
    \label{fig:example}%
\end{figure}
\subsection{Usage}
The code package for \ourmethodabbrev\ is available on github at \url{https://github.com/daeveraert/gradient-information-optimization}. An example use in 2D (the self-consistency test, in this example):


\begin{lstlisting}[language=Python,numbers=none]
from GIO import GIOKL
import numpy as np
import jax.numpy as jnp
import matplotlib.pyplot as plt

# Create some data
def getX():
    mean = [3,4]
    cov = [[0.5,0],[0,0.5]]
    np.random.seed(1)
    x, y = np.random.multivariate_normal(mean, cov, 100).T
    return jnp.array([[x[i],y[i]] for i in range(len(x))])

def getXTest():
    mean = [3,4]
    cov = [[0.5,0],[0,0.5]]
    np.random.seed(5)
    x, y = np.random.multivariate_normal(mean, cov, 100).T
    return jnp.array([[x[i],y[i]] for i in range(len(x))])

X = getX()
X_test = getXTest()

# Initialize class
gio_kl = GIOKL.GIOKL(uniform_low=0, uniform_high=8, \
                     uniform_start_size=100, dim=2)

# Perform the Algorithm
W, kl_divs, _ = gio_kl.fit(X_test, X, normalize=False)
W = W[100:] # Remove the uniform start

# Plot results
plt.plot(kl_divs)
plt.title("KL Divergence vs. Iterations")
plt.xlabel("Iterations")
plt.ylabel("KL Divergence")
plt.show()
plt.clf()
plt.scatter([each[0] for each in W], [each[1] for each in W], \
            label="Selected Data")
plt.scatter([each[0] for each in X], [each[1] for each in X], \
            label="Target Data")
plt.title("Target Data and Selected Data")
plt.xlabel("Dimension 1")
plt.ylabel("Dimension 2")
plt.legend()
plt.show()
\end{lstlisting}

The above code will print the KL divergence at each iteration and produce the plots in \figref{fig:example}.

A complex example using the quantization-explosion technique on big data with Spark:
\begin{lstlisting}[language=Python,numbers=none]
from GIO import GIOKL
import jax.numpy as jnp
import matplotlib.pyplot as plt
import pyspark.sql.functions as F

# Initialize class
gio_kl = GIOKL.GIOKL(uniform_low=-1, uniform_high=1, uniform_start_size=20, dim=768)

# Read data
train_df, target_df = gio_kl.read_data_from_csv(PATH_TRAIN, PATH_TARGET)

# Quantize data
model_train, model_X, transform_train, transformed_X = \ 
                gio_kl.quantize(train, target, k=1500)

X = jnp.array(model_X.clusterCenters())
train = jnp.array(model_train.clusterCenters())
data = [(i, each.tolist()) for i, each in \ 
        enumerate(model_train.clusterCenters())]
centroids_df = gio_kl.spark.createDataFrame(data=data,\ 
                                            schema=["id", "centroid"])

# Perform the Algorithm
W, kl_divs, _ = gio_kl.fit(train, X, max_iter=300, k=5, \
                  stop_criterion="sequential_increase_tolerance",\ 
                  v_init="jump",\ 
                  lr=0.01)

# Explode back to original data and write resulting data
full_selections_df = gio_kl.explode(W, centroids_df, transform_train)
full_selections_df.select(F.col("_c0"), F.col("_c1")).write \
                  .option("delimiter", "\t").csv(OUTPUT_PATH)

# Plot results
plt.plot(kl_divs)
plt.title("KL Divergence vs. Iterations")
plt.xlabel("Iterations")
plt.ylabel("KL Divergence")
plt.show()
\end{lstlisting}

The main function, \verb|GIOKL.fit|, takes the following arguments:

\begin{itemize}[leftmargin=1em]
\item \verb|train|: training data as a jnp array (jnp is almost identical to numpy) [M, D] shape
\item  \verb|X|: target data as a jnp array [N, D] shape
\item  \verb|D|: initial data as a jnp array, default None. Use None to initialize from 0 (uniform) or a subset of training data
\item  \verb|k|: kth nearest neighbor to use in the KL divergence estimation, default 5
\item  \verb|max_iter|: maximum iterations for the algorithm. One iteration adds one point (cluster)
\item  \verb|stop_criterion|: a string for the stopping criterion, one of the following: \textquotesingle increase\textquotesingle , \textquotesingle max\_resets\textquotesingle , \textquotesingle min\_difference\textquotesingle , \textquotesingle sequential\_increase\_tolerance\textquotesingle , \textquotesingle min\_kl\textquotesingle , \textquotesingle data\_size\textquotesingle . Default is \textquotesingle increase\textquotesingle 
\begin{itemize}
\item  \verb|min_difference|: the minimum difference between prior and current KL divergence for \textquotesingle min\_difference\textquotesingle  stop criterion only. Default is 0
    \item  \verb|resets_allowed|: whether if KL divergence increases, resetting G to the full train is allowed (allows the algorithm to pick duplicates). Must be set to true if the stop criterion is \textquotesingle max\_resets\textquotesingle . Default is False
    \item  \verb|max_resets|: the number of resets allowed for the \textquotesingle max\_resets\textquotesingle\  stop criterion only (a reset resets G to the full train set and allows the algorithm to pick duplicates). Default is 2
    \item  \verb|max_data_size|: the maximum size of data to be selected for the \textquotesingle data\_size\textquotesingle\  stop criterion only, as a percentage (of total data) between 0 and 1. Default is 1
    \item  \verb|min_kl|: the minimum kl divergence for the \textquotesingle min\_kl\textquotesingle  stop criterion only. Default is 0
    \item  \verb|max_sequential_increases|: the maximum number of sequential KL divergence increases for the \textquotesingle sequential\_increase\_tolerance\textquotesingle\  stop criterion only. Default is 3
\end{itemize}
\item  \verb|random_init_pct|: the percent of training data to initialize the algorithm from. Default is 0
\item  \verb|random_restart_prob|: probability at any given iteration to extend the gradient descent iterations by 3x, to find potentially better extrema. Higher values come at the cost of efficiency. Default is 0
\item  \verb|scale_factor|: factor to scale the gradient by, or \textquotesingle auto\textquotesingle . Default is \textquotesingle auto\textquotesingle , which is recommended
\item  \verb|v_init|: how to initialize v in gradients descent, one of the following: \textquotesingle mean\textquotesingle , \textquotesingle prev\_opt\textquotesingle , \textquotesingle jump\textquotesingle . Default is \textquotesingle mean\textquotesingle 
\item  \verb|grad_desc_iter|: the number of iterations to use in gradient descent. Default is 50
\item  \verb|discard_nearest_for_xy|: discard nearest in the xy calculation of KL divergence, for use when X and the train set are the same, comes at the cost of efficiency. Default is False
\item  \verb|normalize|: Whether to normalize the uniform start values. Use when the values of X and Train are normalized, as when using embeddings generated by MPNet or MiniLM, for example. Default is True
\item  \verb|lr|: Learning rate for gradient descent. Default is 0.01
\end{itemize}

\subsection{Stopping Criteria}\label{app:algorithm:stop_crit}
We discuss several possible stopping criteria implemented in the code package and outline the pros and cons of each.
\begin{itemize}[leftmargin=2em]
\item \textbf{Strict}: The strictest stopping criterion (\verb|stopping_criterion="increase"|) is to stop immediately when the KL divergence increases from the previous value. This stopping criterion makes the most theoretical sense, as a KL divergence increase indicates the point being added does not add any information about the target X. This is the criterion we use for our text experiments. In practice however, it may be wise to allow some tolerance of KL divergence increase, which we discuss under the Sequential Increase Tolerance item.
\item \textbf{Convergence Tolerance}: Closely related to the strictest stopping criterion is to specify a tolerance of the difference between the prior and current KL divergence and terminate when the difference falls below this tolerance (\verb|stopping_criterion="min_difference"|). The tolerance is similar to tolerance arguments for gradient descent algorithms and is designed to heuristically identify when the algorithm has converged and is no longer decreasing by more than the specified amount.
\item \textbf{Minimum KL Divergence}: The algorithm can terminate when it reaches a pre-determined KL divergence (\verb|stopping_criterion="min_kl"|). In practice, however, it may be difficult to know beforehand what a good value of KL divergence may be, particularly in high dimensions.
\item \textbf{Maximum Data Size}: The algorithm can terminate when it reaches a certain data size. This is a particularly useful stopping criterion when data size/resource constraints are the primary reason for data selection (\verb|stopping_criterion="max_data_size"|). However, as it does not use intrinsic properties of the data that was selected (e.g. Kl divergence), it is the least theoretically-motivated stopping criterion. We use this stopping criterion for the FashionMNIST task in \secref{sec:fashionmnist} (see \appref{app:fashionmnist} for more details)
\item \textbf{Sequential Increase Tolerance}: Instead of stopping when the KL divergence increases as in the strict criterion, we can allow the algorithm a certain amount of sequential increases in KL divergence before terminated. For example, a value of 3 would tell the algorithm to terminate after 3 consecutive points added each increased the KL divergence. If the KL divergence increases, then decreases again, the algorithm can continue. This criterion can allow the algorithm to attain better minima. Primarily, the space is non-convex, and allowing temporary increases in KL divergence can enable the algorithm to get over certain "humps" and descend into more ideal space. However, it could potentially lead to divergence in KL divergence, and also add many suboptimal points if KL divergence increases happen frequently or the tolerance is set too high.

\end{itemize}
\ourmethodabbrev\ restricts points chosen to only be unique points. However, in some cases, allowing the algorithm to choose duplicate points can be beneficial. Intuitively, this would allow the algorithm to weight points in certain areas of the space it feels are most important. In practice, we can let the algorithm run until it reaches the chosen stopping criterion, then reset $G$ and allow the algorithm to pick again. We use this concept in our spelling correction experiments. In the code package, these options are given by the arguments \verb|resets_allowed| (True or False) and  \verb|max_resets|, which specifies how many resets are allowed until full termination.

\subsection{Initial \textbf{v} in Gradient Descent}
We discuss the initial value of $\mathbf{v}$ in the gradient descent implemented in the code package and outline the pros and cons of each.
\begin{itemize}[leftmargin=2em]
\item \textbf{Mean}: The simplest initialization for $\mathbf{v}_{0}$ is to set it equal to the mean of the target $X$ at each iteration of \ourmethodabbrev\ (\verb|v_init="mean"|). This is motivated by the fact that the optimal $\mathbf{v}_{opt}$ lies somewhere in the space of $X$, and taking the mean will make $\mathbf{v}_{0}$ close to $X$ and is a good place to start searching. In unimodal symmetric distributions, the mean will essentially be the optimal point to add. In multimodal distributions, the mean provides a neutral starting point from gradient descent by which it can choose which mode to strike out for, based on the gradient. However, particularly in high dimensions, the mean may be far from the optimal point. Additionally, in scenarios where the modes are far apart and the distribution (and therefore the mean) is skewed to one section, starting from the skewed mean can miss the other sections of the space.
\item \textbf{Previous Optimal Point}: We can also start $\mathbf{v}_{0}=\mathbf{v}_{opt}$, the previous optimal point found by gradient descent (\verb|v_init="prev_opt"|). This is motivated by the fact that adding a new point is unlikely to alter the distribution much, and therefore the next optimal point is likely close to the previous optimal point. This is the setting we use in our text tasks. However in multimodal distributions with modes that are far apart, this may settle for only one or a few modes and not explore the rest of the space, as we are biasing the gradient descent to already-explored areas of the space. At the first iteration we can start from the mean.
\item \textbf{Random Jump}: Instead of a deterministic starting point, we can set $\mathbf{v}_{0}$ to a random value from the target $X$ at each point in the iteration. This setting will explore various parts of the space, and algorithms better suited to non-convex spaces (e.g. particle swarm optimization \citep{pso}) also utilize stochastic methods as an instrument to achieve better convergence. Therefore, this setting can yield more diverse values and potentially better optima, with the risk of not being able to fully explore a particular area of the space. We use this technique in the FashionMNIST task, as in that task we know that the 10 classes of images will be multimodally distributed and the modes relatively delineated from each other, a setup which could cause issues with the previous two settings. 
\end{itemize}
A combination of the methods, and additional methods, could perform even better and we leave this exploration to future work.

\section{Experiments}\label{app:experiments}
We provide the details of our experiments. All code to replicate experiments is included in the github repository (https://github.com/daeveraert/gradient-information-optimization), see \appref{app:algorithm} for details on usage.

\subsection{Analytic Checks}\label{app:analytics}
\begin{figure}[tp]
\centering
\includegraphics[width=\textwidth]{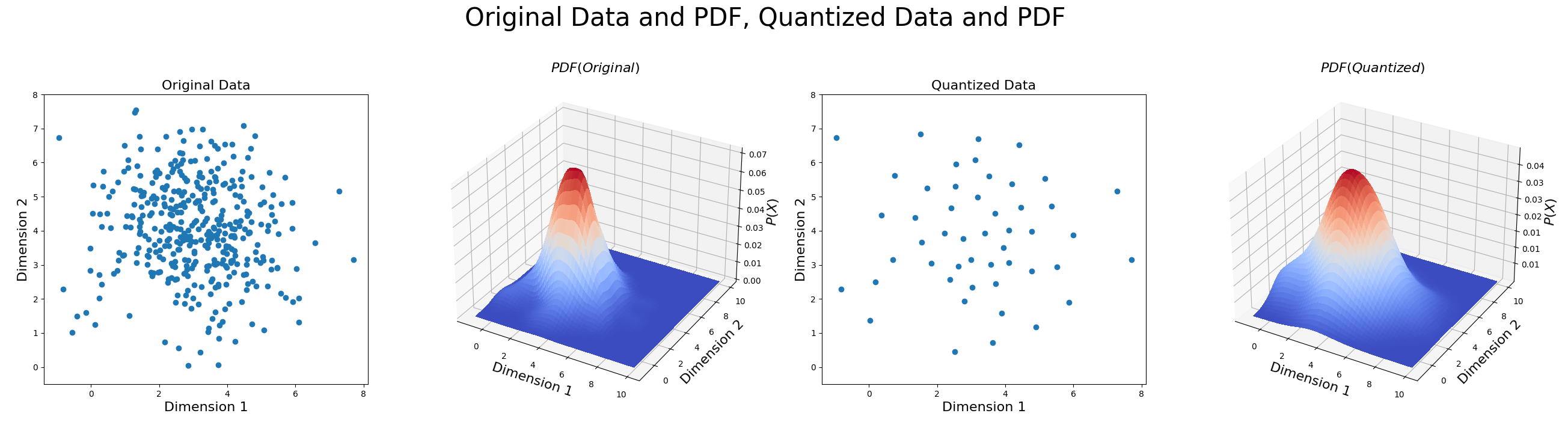}
\caption{Visualization of quantization consistency. Left to right: Original data and kernel density-estimated PDF, Quantized data and kernel density-estimated PDF. The distributions are very close between the quantized space and original space.}
\label{fig:quant-consist}
\end{figure}

We provide the sample code files to run the analytic checks in our code package at the path gradient-information-optimization/experiments/checks/:
\begin{itemize}[leftmargin=2em]
\item \textbf{Self-Consistency}: \url{self_consistency.py}
\item \textbf{Negative-Consistency}: \url{negative_consistency.py}
\item \textbf{Quantization-Consistency}: \url{quantization_consistency.py}
\end{itemize}
None of these code files require arguments, usage is \verb|python FILE.py|. \figref{fig:quant-consist} shows a visualization of the quantization consistency by plotting the estimated probability density functions of the original and quantized space. \figref{fig:example} shows a visualization of self consistency by plotting the target and selected data.

\subsection{WMT14}\label{app:wmt}
This experiment aims to demonstrate using \ourmethodabbrev\ on a well-known dataset with a well-known setup, and shows \ourmethodabbrev\ can achieve similar and even superior performance with significantly less data. See \secref{sec:wmt14} for details on the results of this experiment. We follow Fairseq's recommendations for replicating \citet{vaswani2017transformer} setup.

\subsubsection{Data}\label{app:wmt:data}
We load and preprocess the data using \url{https://github.com/facebookresearch/fairseq/blob/main/examples/translation/prepare-wmt14en2de.sh} for EN-DE and \url{https://github.com/facebookresearch/fairseq/blob/main/examples/translation/prepare-wmt14en2fr.sh} for EN-FR. These scripts handle downloading, preprocessing, tokenizing and preparing the train, valid and test data. For our experiments, we combine the train and valid data from this to use in all of ours, competitive methods and random setups, and recut new validation sets for each experiment.

For the dev sets as our target $X$, we collect the dev sets for EN-FR and EN-DE from WMT08-WMT13 from the provided official dev sets of \url{https://www.statmt.org/wmt14/translation-task.html} and preprocess them using the same techniques and tokenizers given by the Fairseq processing scripts above. We then shuffle and cut 3k from the resulting combined dev sets to be used as our Dev Test set, and keep the remaining 12k as our target $X$.

\subsubsection{\ourmethodabbrev}
We first generate embeddings for the train and dev sets using MPNet-Base-V2 \citep{song2020mpnet} on the input side of the data, using the following code from the code package:
\begin{lstlisting}[language=python,numbers=none]
from GIO import generate_text_embeddings
gen_embeds = generate_text_embeddings.GenerateEmbeddings("all-mpnet-base-v2")
gen_embeds.generate_embeddings(PATH_2_INPUT_SIDE_OF_DATA, OUTPUT_PATH)
\end{lstlisting}
We then paste the original "input \textbackslash t output" with the embeddings file to create a file of the format "input \textbackslash t output \textbackslash t embedding" and use this as our CSV in \ourmethodabbrev. We used AWS p3dn.24xlarge machines to generate the embeddings, which have 8 NVIDIA Tesla V100 GPUs and as a benchmark, takes roughly 4 hours to generate 15M embeddings (therefore approx. 8 hours for EN-FR and approx. 1 hour for EN-DE). This process is highly parallelizable for speed, and additionally, more lightweight models like MiniLM (which takes roughly half the time) and even non-neural methods like Sentence2Vec can be used under speed and resource constraints.
Across all initializations, we use the following parameters:
\begin{itemize}[leftmargin=2em]
\item \textbf{K}: 1500
\item \textbf{k in $\mathbf{\hat D_{KL}}$}: 5
\item \textbf{Max iterations}: 1000
\item \textbf{Stopping Criterion}: increase
\item \textbf{v\_init}: prev\_opt
\item \textbf{Resets Allowed}: false
\item \textbf{Iterations in Gradient Descent}: 50
\item \textbf{Gradient Descent Learning Rate}: 0.01
\end{itemize}

For 0\% initialization, we use a uniform start of 20 points, spread from -1 to 1 with 768 dimensions. For 25\% and 50\% initialization, we start with a random subsample of 375 and 750 clusters respectively, out of the 1500. The following is the quantization and main method signatures for 0\% initialization:

\begin{lstlisting}[language=Python,numbers=none]
...
# Initialize class
gio_kl = GIOKL.GIOKL()

# Read data
train, target = gio_kl.read_data_from_csv(PATH_TRAIN, PATH_TARGET)

# Quantize data
model_train, model_X, transform_train, transformed_X = \ 
                gio_kl.quantize(train, target, k=1500)

X = jnp.array(model_X.clusterCenters())
train = jnp.array(model_train.clusterCenters())
data = [(i, each.tolist()) for i, each in \ 
        enumerate(model_train.clusterCenters())]
centroids_df = gio_kl.spark.createDataFrame(data=data,\ 
                                            schema=["id", "centroid"])

# Perform the Algorithm
W, kl_divs, _ = gio_kl.fit(train, X, max_iter=1000, k=5, \
                            stop_criterion="increase",\ 
                            v_init="prev_opt",\ 
                            resets_allowed=False, grad_desc_iter=50,\ 
                            lr=0.01)
W = W[20:] # Remove uniform start

# Explode back to original data and write resulting data
full_selections_df = gio_kl.explode(W, transform_train, centroids_df)
full_selections_df.select(F.col("_c0"), F.col("_c1")).write \
                  .option("delimiter", "\t").csv(OUTPUT_PATH)
\end{lstlisting}
For 0\% initialization, remove the variable D (both from when it is assigned to a random sample and from the main \verb|fit| function call). For 50\% initialization, increase 375 to 750.

\begin{figure}[tp]
\centering
\includegraphics[width=0.83\textwidth]{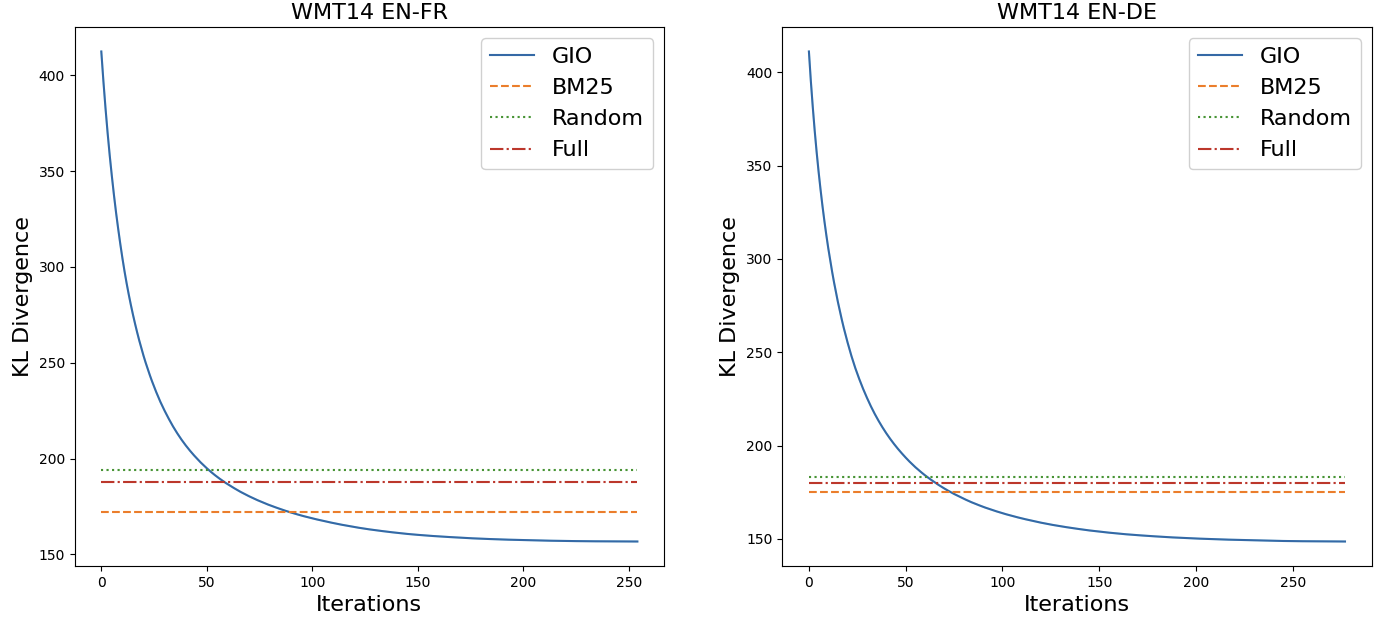}
\caption{\ourmethodabbrev\ surpasses the KL divergence of the most competitive baseline, BM25, at around 80 iterations and ends significantly lower}
\label{fig:wmtkl}
\end{figure}

We performed this algorithm in a Spark environment on a cluster of 30 AWS r4.8xlarge CPU-only instances. For EN-DE, the algorithm took 40min end-to-end. For EN-FR, the algorithm took 50min end-to-end. We saved the initial K means quantized data for reuse in the different initialization states and during experimentation, which saves 10min for DE and 20min for FR on subsequent runs of the algorithm. In \figref{fig:wmtkl}, we show the KL divergence graphs over the iterations for EN-FR and EN-DE at 0\% initialization, and include lines indicating the divergence for BM25, Pruning, Submodular Optimization, Random and Full baselines.

\subsubsection{Baselines}\label{app:bm25}
For the random baselines, after running \ourmethodabbrev, we take a random sample of the same size as that chosen by \ourmethodabbrev. For BM25, we also select sample of the same size as that chosen by \ourmethodabbrev. \citet{yao22c} provide a Github implementation of their data selection technique at \url{https://github.com/yaoxingcheng/TLM} \citep{yaogit}. We use the \verb|data_selection.py|, for example:

\begin{lstlisting}[numbers=none]
python src/data_selection.py \
    --index_name example_bm25_source \
    --source_file SOURCE.csv \
    --target_file TARGET.csv \
    --output_dir ./example_data \
    --output_name selected.csv \
    --top_k K
\end{lstlisting}
The source file is the (input only) data we can select from, $G$, and the target file is the (input only) $X$, in this case the collected dev sets. Top K is the parameter that tells BM25 how much data to retrieve for each data point in $X$. Since BM25 can output a score of 0 and discard data, we cannot calculate K as $K = \frac{|G|}{|X|}$, and typically, we need to overestimate K to collect enough data to match the desired size. We also need to pre and post process our data to conform to their code, which we address and provide scripts for.

\paragraph{Preproccessing.} Their implementation requires a CSV with columns text,id, but data given to us with the WMT14 scripts is in plain text format with no ID. We provide a script at \url{gradient-information-optimization/experiments/bm25_scripts/make_csv.py} which transforms a single-column file of inputs into the format needed by the \citet{yao22c} code:
\begin{lstlisting}[numbers=none]
python make_csv.py INPUT_FILE TO_OUTPUT_FILE
\end{lstlisting}

\paragraph{Postprocessing.} Their implementation will output a CSV with the format chosen\_data,input ID,rank. The rank is the value of "K" for which this data was retrieved. For example, if this was the 5th most relevant to a data point, then the rank will read 5. As addressed, we need to overestimate K to collect enough samples. Inevitably, we end up with too many samples, and need to trim down the data by removing the least relevant retrievals. We provide a script at \url{gradient-information-optimization/experiments/bm25_scripts/get_less.py} which takes in the file output by their code, an output path, and a value of rank to filter; we keep the values under the specified rank value, and discard any over. This is a trial-and-error process, repeatedly trying different values of K until we get the desired data size. Usage is:
\begin{lstlisting}[numbers=none]
python get_less.py INPUT_FILE TO_OUTPUT_FILE K_FILTER_VALUE
\end{lstlisting}
Finally, we need to recover the input-output pairs, as the output file only has input-side data. We provide a script at \url{gradient-information-optimization/experiments/bm25_scripts/get_pairs.py} that retrieves the pairs from input side given the input data and original data of the format \verb|input \t output|:
\begin{lstlisting}[numbers=none]
python get_pairs.py ORIGINAL_DATA INPUT_FILE TO_OUTPUT_FILE
\end{lstlisting}

For submodular optimization, we use the same cluster centers as \ourmethodabbrev, the LazyGreedy optimizer and facility location mutual information function, and run to have the same data size as the \ourmethodabbrev-selected data. We provide a script at \url{gradient-information-optimization/experiments/submod/submod.py} which takes three inputs and an output argument. The first input should be a parquet file with one column of array<double> for the data and the second input should be the same for the target $X$. The third input should be the threshold data size, determined by $\frac{\text{Desired Data Size}}{\text{Total Data Size}} \cdot \text{Cluster Size}$, and the output should also be a parquet file with a column array<double>.
\begin{lstlisting}[numbers=none]
python submod.py DATA_CLUSTERS X_CLUSTERS THRESHOLD TO_OUTPUT_FOLDER
\end{lstlisting}

For self-pruning,  we use the same cluster centers as \ourmethodabbrev. We adapt their technique in \url{gradient-information-optimization/experiments/pruning/prune.py} which takes three inputs and one output argument. The first input should be a parquet file of the cluster centers and the second input should be the transformed dataframe from Spark's KMeans. The third input should be the threshold, which is calculated by the formula $\frac{\text{Desired Data Size}}{\text{Cluster Size}}$. The output should be output folder.
\begin{lstlisting}[numbers=none]
python prune.py INPUT_CLUSTERS TRANSFORMED_DF THRESHOLD TO_OUTPUT_FOLDER
\end{lstlisting}

\subsubsection{Training}
We preprocess and train using Fairseq \citep{fairseq}. We cut 3k pairs from the data for validation and keep the rest as training data. We use the subword-nmt package to apply BPE \cite{sennrich-etal-2016-neural, subwordgit} using the codes already computed and provided by the Fairseq scripts to download data (\appref{app:wmt:data}). We then use \verb|fairseq-preprocess| to preprocess the data into Fairseq-readable format. We then follow the \citet{vaswani2017transformer} setup and train for 300k iterations, saving every 15k iterations. We use Fairseq's pre-built \verb|transformer_vaswani_wmt_en_fr_big| architecure for EN-FR and \verb|transformer_vaswani_wmt_en_fr_big| for EN-DE. The pre-built architectures preset essentially all of the model parameters described by \citet{vaswani2017transformer}. We followed the recommendation in \url{https://github.com/facebookresearch/fairseq/issues/346} to set the LR at 0.0007 for close replication of \citet{vaswani2017transformer} with the Fairseq framework. We varied the max\_tokens parameter to ensure approximately 25k tokens per batch, as recommended in the paper \citep{vaswani2017transformer}.

\begin{center}
    \textbf{EN-FR CLI}
\end{center}
\begin{lstlisting}[numbers=none]
$(which fairseq-train) DATA_PREPROCESSED \                                      
    --arch transformer_vaswani_wmt_en_fr_big \
    --share-all-embeddings \
    --optimizer adam --lr 0.0007 \
    --dropout 0.1 \
    --max-tokens 3512 \
    --lr-scheduler inverse_sqrt --weight-decay 0.0 \
    --criterion label_smoothed_cross_entropy --label-smoothing 0.1 \
    --fp16 --save-dir SAVE_DIR --save-interval-updates 15000 \
    --max-update 300000 --clip-norm 0.0 --warmup-updates 4000 \
    --warmup-init-lr 1e-07 --min-lr 1e-09 --adam-betas '(0.9,0.98)'
\end{lstlisting}
\begin{center}
    \textbf{EN-DE CLI}
\end{center}
\begin{lstlisting}[numbers=none]
$(which fairseq-train) DATA_PREPROCESSED \                                      
    --arch transformer_vaswani_wmt_en_de_big \
    --share-all-embeddings \
    --optimizer adam --lr 0.0007 \
    --dropout 0.3 \
    --max-tokens 3512 \
    --lr-scheduler inverse_sqrt --weight-decay 0.0 \
    --criterion label_smoothed_cross_entropy --label-smoothing 0.1 \
    --fp16 --save-dir SAVE_DIR --save-interval-updates 15000 \
    --max-update 300000 --clip-norm 0.0 --warmup-updates 4000 \
    --warmup-init-lr 1e-07 --min-lr 1e-09 --adam-betas '(0.9,0.98)'
\end{lstlisting}
We trained on one AWS p3dn.24xlarge, which has 8 NVIDIA Tesla V100 GPUs and takes about 10 hours to train the full 300k iterations.

\subsubsection{Testing}
We use the script provided by Fairseq moderators in \url{https://gist.github.com/myleott/da0ea3ce8ee7582b034b9711698d5c16} \citep{ottbleu} for the evaluation process. In order to generate the predictions on the preprocessed test data from the best model by validation loss, we use the checkpoint\_best.pt model produced by Fairseq and the command:
\begin{lstlisting}[numbers=none]
fairseq-generate TEST_DATA --path MODEL --beam 4 --lenpen 0.6 \
                           --remove-bpe > outfile 
\end{lstlisting}
The outfile then gets fed into the script provided by the Fairseq moderators to give the BLEU score.

\subsection{Robustness}
The robustness experiments follow a nearly identical setup to the above WMT14 setup. We outline the differences:
\paragraph{MiniLM.} To use MiniLM to generate the embeddings for \ourmethodabbrev, we replace \verb|"all-mpnet-base-v2"| with \verb|"all-MiniLM-L12-v1"| in the GenerateEmbeddings class instantiation:
\begin{lstlisting}[language=python,numbers=none]
from GIO import generate_text_embeddings as gte
gen_embeds = gte.GenerateEmbeddings("all-MiniLM-L12-v1")
gen_embeds.generate_embeddings(PATH_2_INPUT_SIDE_OF_DATA, OUTPUT_PATH)
\end{lstlisting}
MiniLM takes roughly half the time than MPNet to generate embeddings. On an AWS p3dn.24xlarge machine with 8 NVIDIA Tesla V100 GPUs, it takes roughly 30min for EN-DE and roughly 4 hours for EN-FR. In addition, we alter the instantiation of the GIOKL class to specify dimensions of 384:
\begin{lstlisting}[language=python,numbers=none]
...
gio_kl = GIOKL.GIOKL(dim=384)
...
\end{lstlisting}
All else stays the same as the WMT14 setup.

\paragraph{K=1000 and K=3000.} For the experiments altering the value of K, we reuse all the embeddings generated with MPNet for the main experiments. The only difference is to the quantize method signature with the difference values of K:
\begin{lstlisting}[language=python,numbers=none]
...
model_train , model_X , transformed_train , transformed_X = \
                 gio_kl.quantize(train, target , k=1000) # Or k=3000
...
\end{lstlisting}
All else stays the same as the WMT14 setup. 

When computing the variance from the base setup for data size, we omit the K=1000 data size since (as covered in \secref{sec:robustness}) we need to artificially subsample to keep the selected data size the same as in the base setup. This is due to the coarser grain of K=1000 selecting more data. Interestingly, K=3000 and MiniLM selects roughly the same data as the base setup.

\subsection{Speller}

This experiment aims to demonstrate using \ourmethodabbrev\ on a mix of high and low quality data, and aims to show \ourmethodabbrev\ selects the high quality data. We use the spelling correction domain for this problem, as the process of noising training data to create training pairs makes it easy to create high and low quality synthetic data by varying the noise. We determine the \% of high quality data selected by each method, and also train a model and show spelling performance of our method, random subset, BM25, pruning and submodular optimization and show the results of \ourmethodabbrev-selected data are better than that trained on the full mix of low and high quality data \tabref{table-speller2}. We generally follow the recent work of \citet{neuspell} for this problem.

\subsubsection{Data}
\citet{neuspell} use the 1 Billion Word Corpus and extract roughly 1M pairs for creating synthetic data. In order to show our method at scale, we download the full 1 Billion Word Corpus from \url{https://www.statmt.org/lm-benchmark/} and extract 15M data. We add in the 1M pairs used by \citet{neuspell} and deduplicate on label, which results in roughly 14.7M data. When we have the labels, we need to noise them to create the inputs. \citet{neuspell} implement and use 3 different synthetic data noising methods: random, word and prob. They show "prob" is the best method, followed by "word" and then "random". We use the "prob" method on half the data to create our high quality input-output pairs, and use the "word" method on the remaining half. The "word" method takes in a probability that a word will be replaced by a misspelled version of it contained in a lookup table, if it exists. \citet{neuspell} use a probability of 20\%, but to make very low quality data, we use a high probability of 70\%. They provide a Github implementation of their methods and data at \url{https://github.com/neuspell/neuspell} \citep{neuspellgit}, which we use to noise the data accordingly. We give an example of the quality difference in \tabref{tab:quality-comp}.

\begin{table}[h]
  \caption{Right: High Quality "Prob" data. Left: Low Quality "Word" data. High quality data represents reasonable spelling mistakes, low quality data is barely intelligible. \ourmethodabbrev\ selects 73\% from the high quality data.}
  \centering
  {\begin{tabularx}{\textwidth}{X X}
    \toprule
    \textbf{High Quality "Prob"} & \textbf{Low Quality "Word"}\\
    \midrule
    It is still a big poblem though & It ls sttel a beig probelma thouigh\\
    It apoeared they were not serousay injured , the BBC said & It appered thiy whir nt serioulsly ingerd , whe BBC sayd\\
    Were they aarried on Feb & Were thwey marreid ong Feb\\
    More of us are tuning in to tha irwaves than ever & More th uus spe tuning inf tome vthe airwaves thanx ef\\
    \bottomrule
  \end{tabularx}}\label{tab:quality-comp}

\end{table}
The high quality "prob" data is more reflective of real spelling mistakes, and in fact the "prob" method uses a context matrix to make applicable misspellings in context. The "word" method at 70\%, on the other hand, severely mangles the sentences to the point where we barely make out what they mean, making for very low quality data to train with. After generating half with prob and half with word, we recombine and shuffle to create our training data. For our target $X$ set, \citet{neuspell} provide 40k pairs of real mistakes by humans taken from the BEA grammar correction corpus \citep{bea}, which we use as our ideal target $X$.

\subsubsection{\ourmethodabbrev}
We first generate embeddings for the train and target sets using MPNet-Base-V2 \citep{song2020mpnet} on the input side of the data; see \appref{app:wmt} for an example, we use the same process as in our WMT experiments. On an AWS p3dn.24xlarge machine with 8 NVIDIA Tesla V100 GPUs, it takes about 4 hours to generate the embeddings for the 15M data. As we mention in \secref{sec:speller}, we experiment with a new scheme. Instead of stopping on an increase, we allow the algorithm to reset $G$ once when the KL divergence increases, and pick again until the divergence increases a second time. We hope this will allow the algorithm to pick not only high quality data, but emphasize (by picking duplicates) the points that are also the most relevant for $X$. We use the following parameters in the algorithm:
\begin{itemize}[leftmargin=2em]
\item \textbf{K}: 1500
\item \textbf{k in $\mathbf{\hat D_{KL}}$}: 5
\item \textbf{Max iterations}: 1000
\item \textbf{Stopping Criterion}: increase
\item \textbf{v\_init}: prev\_opt
\item \textbf{Resets Allowed}: True
\item \textbf{Maximum Resets}: 1
\item \textbf{Iterations in Gradient Descent}: 50
\item \textbf{Gradient Descent Learning Rate}: 0.01
\end{itemize}

The method is as follows:

\begin{figure}[tp]
\centering
\includegraphics[width=0.5\textwidth]{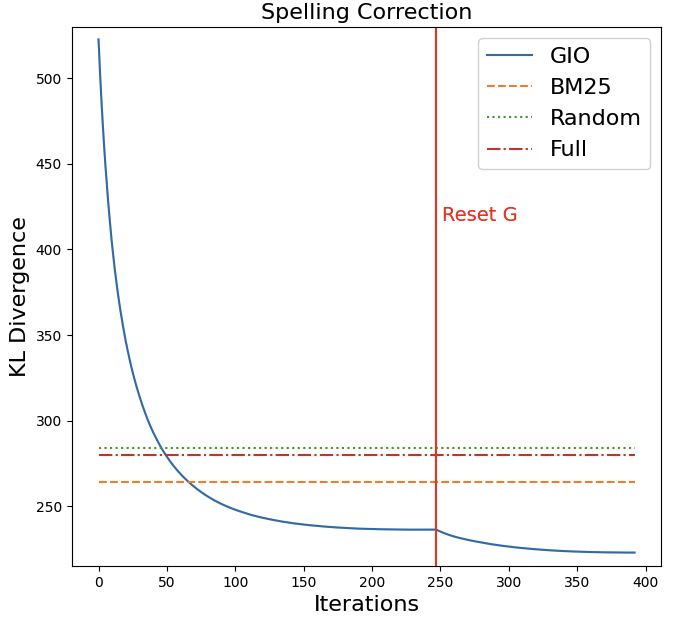}
\caption{ Resetting G enables the algorithm to make further progress reducing KL divergence.}
\label{fig:spellerkl}
\end{figure}

\begin{lstlisting}[language=Python,numbers=none]
...
# Initialize class
gio_kl = GIOKL.GIOKL()

# Read data
train, target = gio_kl.read_data_from_csv(PATH_TRAIN, PATH_TARGET)

# Quantize data
model_train, model_X, transform_train, transformed_X = \ 
                gio_kl.quantize(train, target, k=1500)

X = jnp.array(model_X.clusterCenters())
train = jnp.array(model_train.clusterCenters())
data = [(i, each.tolist()) for i, each in \ 
        enumerate(model_train.clusterCenters())]
centroids_df = gio_kl.spark.createDataFrame(data=data,\ 
                                            schema=["id", "centroid"])

# Perform the Algorithm
W, kl_divs, _ = gio_kl.fit(train, X, max_iter=1000, k=5, \
                            stop_criterion="increase",\ 
                            resets_allowed=True, max_resets=1, \
                            v_init="prev_opt", grad_desc_iter=50,\ 
                            lr=0.01)
W = W[20:] # Remove uniform start

# Explode back to original data and write resulting data
full_selections_df = gio_kl.explode(W, transform_train, centroids_df)
full_selections_df.select(F.col("_c0"), F.col("_c1")).write \
                  .option("delimiter", "\t").csv(OUTPUT_PATH)
\end{lstlisting}
We performed this algorithm in the same Spark environment on a cluster of 30 AWS r4.8xlarge CPU-only instances, which took around 1h15m to complete. In \figref{fig:spellerkl}, we show the KL divergence graphs over the iterations and include lines indicating the divergence for BM25, Pruning, Submodular Optimization, Random and Full baselines. We mark where the algorithm reset $G$; we can see the KL divergence further decreases.

\subsubsection{Baselines}
For the random baseline, after running \ourmethodabbrev, we take a random sample of the same size as that chosen by \ourmethodabbrev. For BM25, submodular optimization and pruning, please see \appref{app:bm25}, as we follow exactly the same process as in our WMT experiments, using the same scripts but on different data. We also train a model on the full data for comparison.

\subsubsection{Training}
We preprocess and train using Fairseq \citep{fairseq}. We cut 3k pairs from the data for validation and keep the rest as training data. We use the subword-nmt package \citep{subwordgit} to learn and apply a 40k BPE from the combined source and target data. We then use \verb|fairseq-preprocess| to preprocess the data into Fairseq-readable format. We train with a seq2seq setup proposed by \citet{spellcorrseq2seq} that learns to map input (noised) to output labels (regular, no noise). We use the BART base architecture \citep{bart} as our seq2seq model, since the denoising objective they used when developing BART is naturally suited to the spelling correction task. We train for 50k iterations and pick the best model checkpoint by validation loss, saving every 5000 iterations. We use 4096 max tokens per batch, the Adam \citep{adam} optimizer with LR of 0.0001, and generally keep the rest of the settings the same as in the BART paper \citep{bart}. For our method and all baselines, we train twice and take the average.

\begin{center}
    \textbf{Speller CLI}
\end{center}
\begin{lstlisting}[numbers=none]
$(which fairseq-train) DATA_PREPROCESSED \                                      
    --arch bart_base \
    --share-all-embeddings \
    --optimizer adam --lr 0.0001 \
    --dropout 0.1 \
    --max-tokens 4096 \
    --lr-scheduler inverse_sqrt \
    --fp16 --save-dir SAVE_DIR --save-interval-updates 5000 \
    --max-update 50000 --clip-norm 0.0 --warmup-updates 4000 \
    --warmup-init-lr 1e-07 --min-lr 1e-09 --adam-betas '(0.9,0.98)'
\end{lstlisting}
We trained on one AWS p3dn.24xlarge, which has 8 NVIDIA Tesla V100 GPUs and takes about 2 hours to train the full 50k iterations.

\subsubsection{Testing}
As test data, \citet{neuspell} provide several sets. We cannot use the BEA 60k they provide as we used 40k from this as our target $X$. Instead, \citet{neuspell} provide an even more challenging set of 4660 intentionally ambiguous pairs curated from BEA60k (the 4660 pairs have no overlap with our $X$), and we use this test set. We tokenize and preprocess this test set with the same codes as our training data. For metrics, \citet{neuspell} report word-level accuracy and word-level correction rate. Word-level accuracy is the fraction of words that the model gets correct, and word-level correction rate is the fraction of words that should be corrected that do get corrected. We also report the F1 score as a balance between these two metrics. In theory it is possible to arbitrarily correct every single word randomly and get 100\% correction rate but 0\% accuracy, or keep every word the same and get near 100\% accuracy but 0\% correction rate. 

An additional complexity arises with our particular use of a seq2seq framework. The models of \citet{neuspell} use a classification objective to classify each input word, which means the output and input sequence will mandatorily have the same number of words. However, this property is not necessarily guaranteed for a seq2seq setup with subword tokenization, and after inferencing our models on test data we see a small number (roughly 1\%) where the input and output number of words do not match. For fairness, when evaluating the models we remove the few  sequences that have a misalignment from the evaluation. In order to generate the predictions on the preprocessed test data from the best model by validation loss, we use the checkpoint\_best.pt model produced by Fairseq and the command:
\begin{lstlisting}[numbers=none]
fairseq-generate TEST_DATA --path MODEL --beam 1 \
                           --remove-bpe > outfile 
grep ^H outfile | cut -f3- > hypotheses
grep ^T outfile | cut -f2- > targets
grep ^S outfile | cut -f2- > sources
\end{lstlisting}
We then feed the hypotheses, targets and sources into a script we provide at \url{gradient-information-optimization/experiments/speller_scripts/calculate_scores.py} which computes the word-level accuracy and correction rate.

\begin{table}[tp]
 \caption{Spelling correction results. Training data size, accuracy/correction rate/F1 scores, \% of high quality data and KL Divergence of the full data, the \ourmethodabbrev-selected data, competitive methods and random subset. \textbf{Bold} is the best score overall. \ourmethodabbrev\ selects 73\% high quality data,  outperforms all other methods and sets a new state-of-the-art on spelling correction.}
  \label{table-speller2}
  \centering
  {\begin{tabular}{l c c c c c c}
    \toprule
    \textbf{System} & \textbf{Train Size} & \textbf{Accuracy} &\textbf{Correction Rate} & \textbf{F1} &  \textbf{\%High Quality} & $\mathbf{\hat D_{KL}}$ \\
    \midrule
    Ours  &  & \textbf{95.9} & \textbf{99.6} & \textbf{97.7} & \textbf{73\%} & \textbf{224}    \\
    BM25 &  & 95.6 & 99.3 & 97.4 & 55\% & 264    \\
    Pruning & \phantom{0}3.6M & 95.7 & 99.6 & 97.6 & 54\% & 241    \\
    Submod. &  & 95.7 & 99.5 & 97.6 & 59\% & 245    \\
    Random  & & \textbf{95.9} & 99.4 & 97.6 & 50\% & 284  \\
    \midrule
    Full & 14.7M & 95.7 & 99.5 & 97.6 & 50\% & 280 \\
    \bottomrule
  \end{tabular}}
\end{table}

\paragraph{Results} \ourmethodabbrev\ outperforms random, all comparative methods, and the full model. \ourmethodabbrev\ outperforms all comparative baselines on accuracy, correction rate and overall F1, and matches random baseline on accuracy and outperforms on correction rate and overall F1. It outperforms the full model by +0.2pps for accuracy and +0.1pps for correction rate, despite using only 24\% of the data. We set a new state-of-the-art in correction rate and overall F1 score on BEA4660 over the best model reported by \citet{neuspell} (+2.3 pps F1 and +7.1 pps correction rate, Table 2 BERT model  in\citet{neuspell}). See \tabref{table-speller2} for details.

\subsection{FashionMNIST}\label{app:fashionmnist}
This experiment aims to demonstrate using \ourmethodabbrev\ can effectively reduce the training data size under a resource constraint without big drops in performance. We use the FashionMNIST image recognition task \citep{fashionmnist} to demonstrate using \ourmethodabbrev\ outside language tasks. Our goal is to reduce the train set size to 25\% of the total size. Experiment results are available at \secref{sec:fashionmnist}.

\subsubsection{Data}
The FashionMNIST task has 60k 28x28x1 images for training and 10k images for testing. We load the images in CSV format from \url{https://www.kaggle.com/datasets/zalando-research/fashionmnist} \citep{fashionmnist_loader}. In our dataloader, we resize our images to 224x224 for Resnet50 using skimage's resize function, normalize each channel, and additionally normalize the values to fall within 0 and 1. The dataloader code is available in the main file we publish at \url{gradient-information-optimization/experiments/fashion_mnist/train.py}. We based the dataloader code on code provided at \url{https://www.kaggle.com/code/pankajj/fashion-mnist-with-pytorch-93-accuracy}. We use the vector format of the images themselves in \ourmethodabbrev. $G$ is the training set, but in this case, $X$ is also the training set. Since our goal is to reduce the training set size, we want to select a subset of $G$ that contains the most information about the whole set, so the target set is the whole set itself.

\subsubsection{\ourmethodabbrev}
In order to read the images using the code package, we do a simple processing to make the image CSV into a string representing an array, the script is available at \url{gradient-information-optimization/experiments/fashion_mnist/csv_to_image.py}; first argument is input, second is output. We also normalize each channel and norm each image. For the algorithm itself, we stop at 25\% data size, and use K=1000. With this problem, we know that there are 10 classes and the images are likely to be clustered relatively distinctly, by class. Therefore, for the initial $\mathbf{v}$ in gradient descent, we choose the random jump scheme, as we will therefore be able to cover more of the space. We use the following parameters in the algorithm:

\begin{itemize}[leftmargin=2em]
\item \textbf{K}: 1000
\item \textbf{k in $\mathbf{\hat D_{KL}}$}: 5
\item \textbf{Max iterations}: 300
\item \textbf{Stopping Criterion}: data\_size
\item \textbf{Maximum Data Size}: 0.25
\item \textbf{v\_init}: jump
\item \textbf{Iterations in Gradient Descent}: 50
\item \textbf{Gradient Descent Learning Rate}: 0.01
\end{itemize}

The method is as follows:

\begin{lstlisting}[language=Python,numbers=none]
...
# Initialize class
gio_kl = GIOKL.GIOKL()

# Read data
train, target = gio_kl.read_data_from_csv(PATH_TRAIN, PATH_TARGET)

# Quantize data
model_train, model_X, transform_train, transformed_X = \ 
                gio_kl.quantize(train, target, k=1000)

X = jnp.array(model_X.clusterCenters())
train = jnp.array(model_train.clusterCenters())
data = [(i, each.tolist()) for i, each in \ 
        enumerate(model_train.clusterCenters())]
centroids_df = gio_kl.spark.createDataFrame(data=data,\ 
                                            schema=["id", "centroid"])

# Perform the Algorithm
W, kl_divs, _ = gio_kl.fit(train, X, max_iter=300, k=5, \
                            stop_criterion="data_size",\ 
                            max_data_size=0.25, v_init="jump", \ 
                            resets_allowed=False, grad_desc_iter=50,\ 
                            lr=0.01)
W = W[20:] # Remove uniform start

# Explode back to original data and write resulting data
full_selections_df = gio_kl.explode(W, transform_train, centroids_df)
full_selections_df.select(F.col("_c0"), F.col("_c1")).write \
                  .option("delimiter", "\t").csv(OUTPUT_PATH)
\end{lstlisting}

\begin{figure}[tp]
\centering
\includegraphics[width=0.5\textwidth]{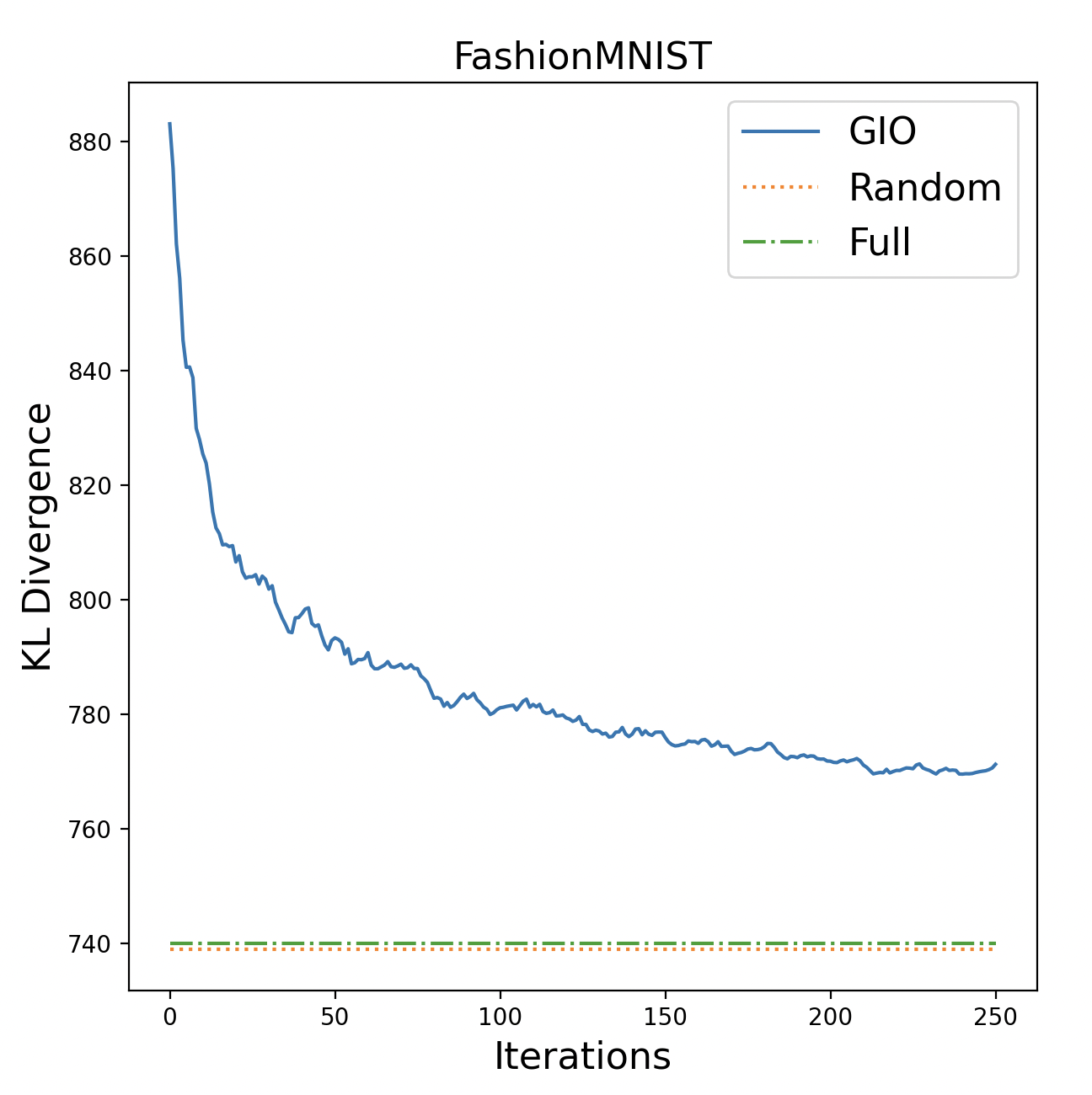}
\caption{\ourmethodabbrev\ approaches the KL divergence of the full data}
\label{fig:fashionkl}
\end{figure}

We performed this algorithm in the same Spark environment on a cluster of 30 AWS r4.8xlarge CPU-only instances, which took around 20m to complete. We show the KL divergence over the iterations. In this case, the minimum KL divergence is not 0, but 739. We explain why the KL divergence is not 0 in \appref{app:0-grad-kl-mod}. The graph of KL divergence is shown in \figref{fig:fashionkl}.
\paragraph{Baselines.} For the random baseline, after running \ourmethodabbrev, we take a random sample of the same size as that chosen by \ourmethodabbrev. We also train a model on full data for comparison, but the goal in this setup is not to match or exceed the full model, but minimize the drop in performance under the resource constraint.

\subsubsection{Training and Testing}
We finetune a pretrained Resnet50 model with the data. For \ourmethodabbrev\ and random, we cut 1,700 as validation and leave the remaining 15,000 (25\%) as our training data. For the full model, we follow this percentage and cut 3,700 as validation and keep 56,300 as the training data. In order to get the images back into the correct CSV format for the dataloader from a string-array format, we provide a script at \url{gradient-information-optimization/experiments/fashion_mnist/image_to_csv.py}; first argument is input, second is output. We edit the architecture of the Resnet50 model on the input layer to take 1 channel instead of 3, and edit the output layer to output to 10 classes. We finetune for 5 epochs with Adam and a small learning rate of 0.00005 so as not to overwrite the existing weights in the pretrained model. We use batch size 100 and take the best checkpoint by validation loss and report the test score on this. The code to read and train a model is available at \url{gradient-information-optimization/experiments/fashion_mnist/train.py}. We based this training script on code provided at \url{https://github.com/kjamithash/Pytorch_DeepLearning_Experiments/blob/master/FashionMNIST_ResNet_TransferLearning.ipynb} \citep{resnet_transfer}. Usage:
\begin{lstlisting}[numbers=none]
python train.py TRAIN_CSV TEST_CSV VALID_CSV
\end{lstlisting}
For testing, the above code will output the test accuracy at each epoch. We train on one AWS p3dn.24xlarge machine, which has 8 NVIDIA Tesla V100 GPUs and takes around 15min to train.

\section{Further FashionMNIST}

\subsection{\ourmethodabbrev\ KL Divergence vs. Random KL Divergence}

It is interesting that despite achieving a higher KL divergence than the random baseline, \ourmethodabbrev\ performs better. We hypothesize that this is due to the fact that \ourmethodabbrev\ is optimizing explicitly for choosing images that add the most information about the whole, i.e., discarding outlier data and focusing on "core" data under the size constraint, which leads to higher KL divergence than random but better performance, as the model will be less confused. For example, in the "bag" category, conceivably a good portion of the bags look very similar, but there is also a portion that are more strange. Random would choose both the core portion of bags that looks similar but also include some of the stranger examples. \ourmethodabbrev, on the other hand, will focus almost exclusively on choosing the core portion, as that is the set that adds the most information, or best represents, the bag category overall. In short, outliers are less important to know under a size constraint, and the core is more important to get right, which \ourmethodabbrev\ chooses. We validate this hypothesis in three ways.

\begin{figure}[tp]
    \centering
    \subfloat{{\includegraphics[width=0.45\textwidth]{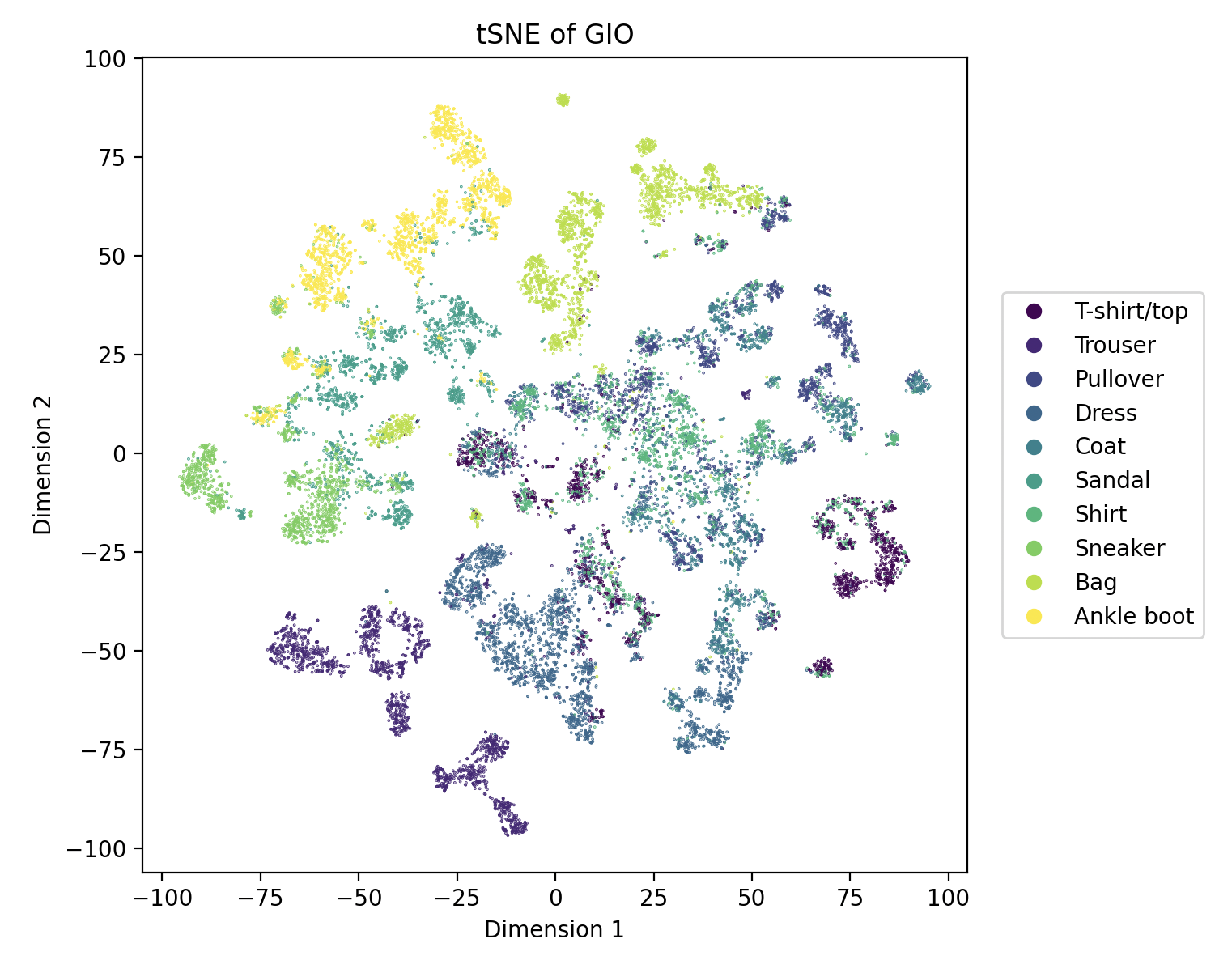} }}
    \subfloat{{\includegraphics[width=0.45\textwidth]{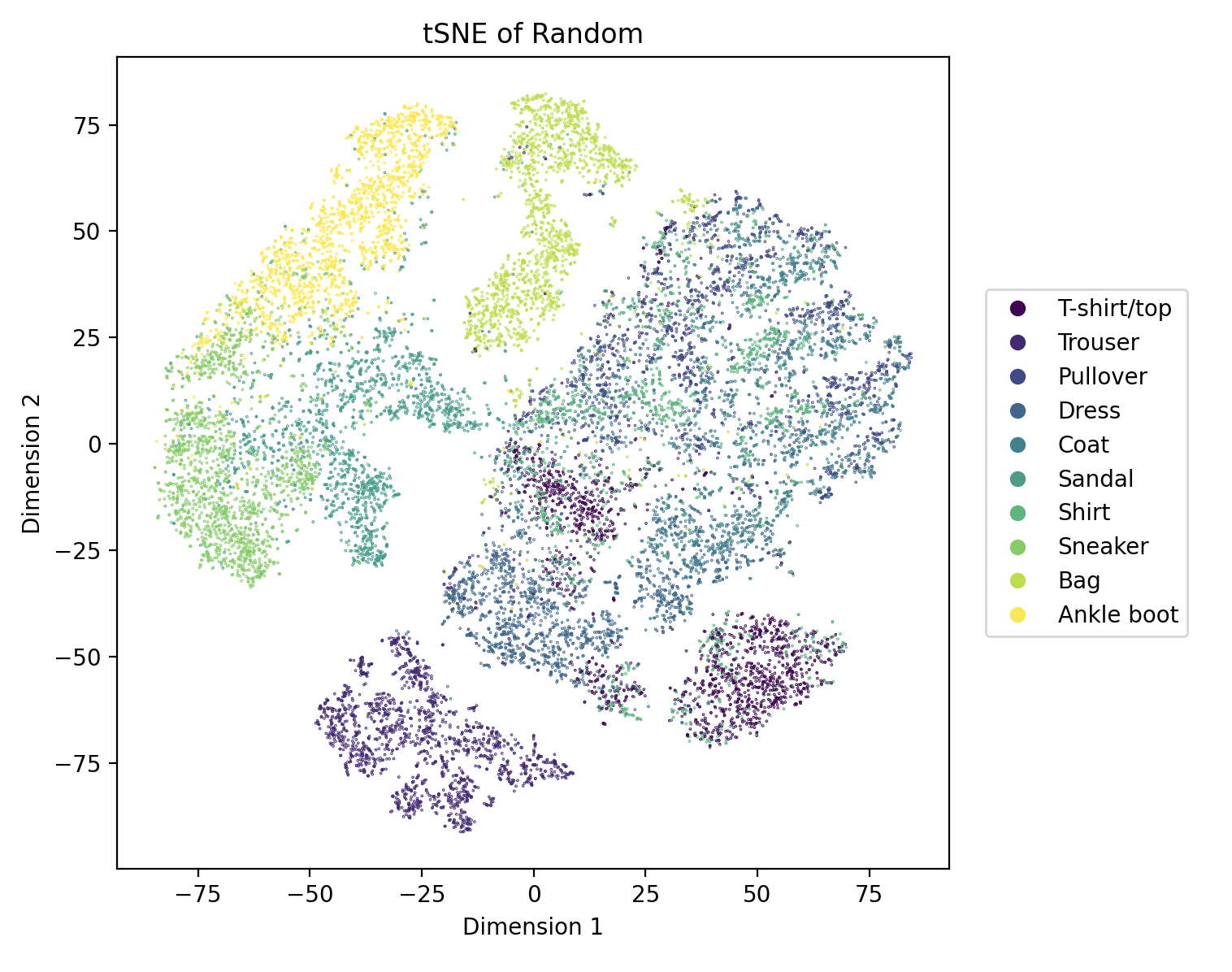} }}
    \caption{Left: tSNE of \ourmethodabbrev, Right: tSNE of Random. \ourmethodabbrev\ is more separable and dense in each class, indicating \ourmethodabbrev\ chooses more core data and fewer outliers}%
    \label{fig:tsne}
\end{figure}

\begin{figure}[tp]
    \centering
    \subfloat{{\includegraphics[width=0.45\textwidth]{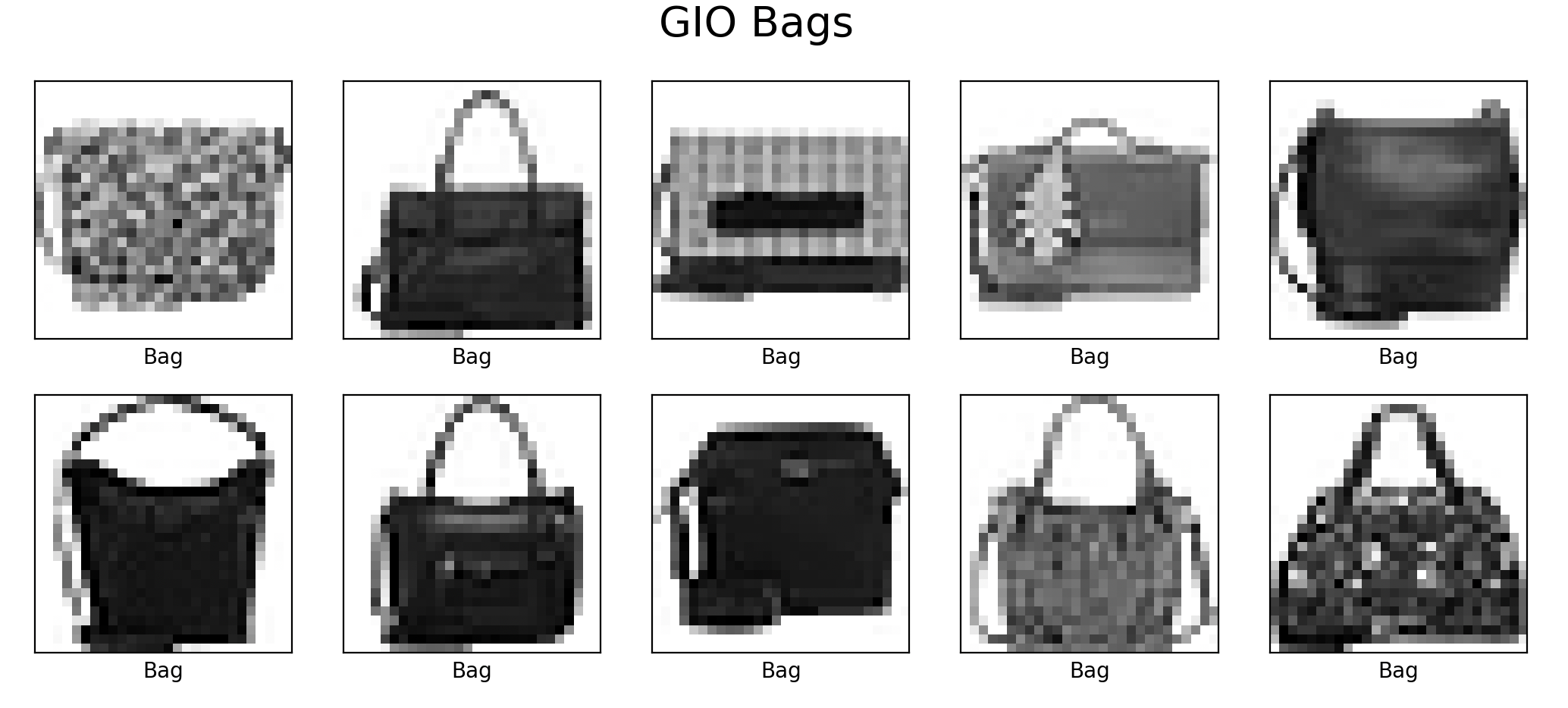} } \hspace{3em}}
    \subfloat{{\includegraphics[width=0.45\textwidth]{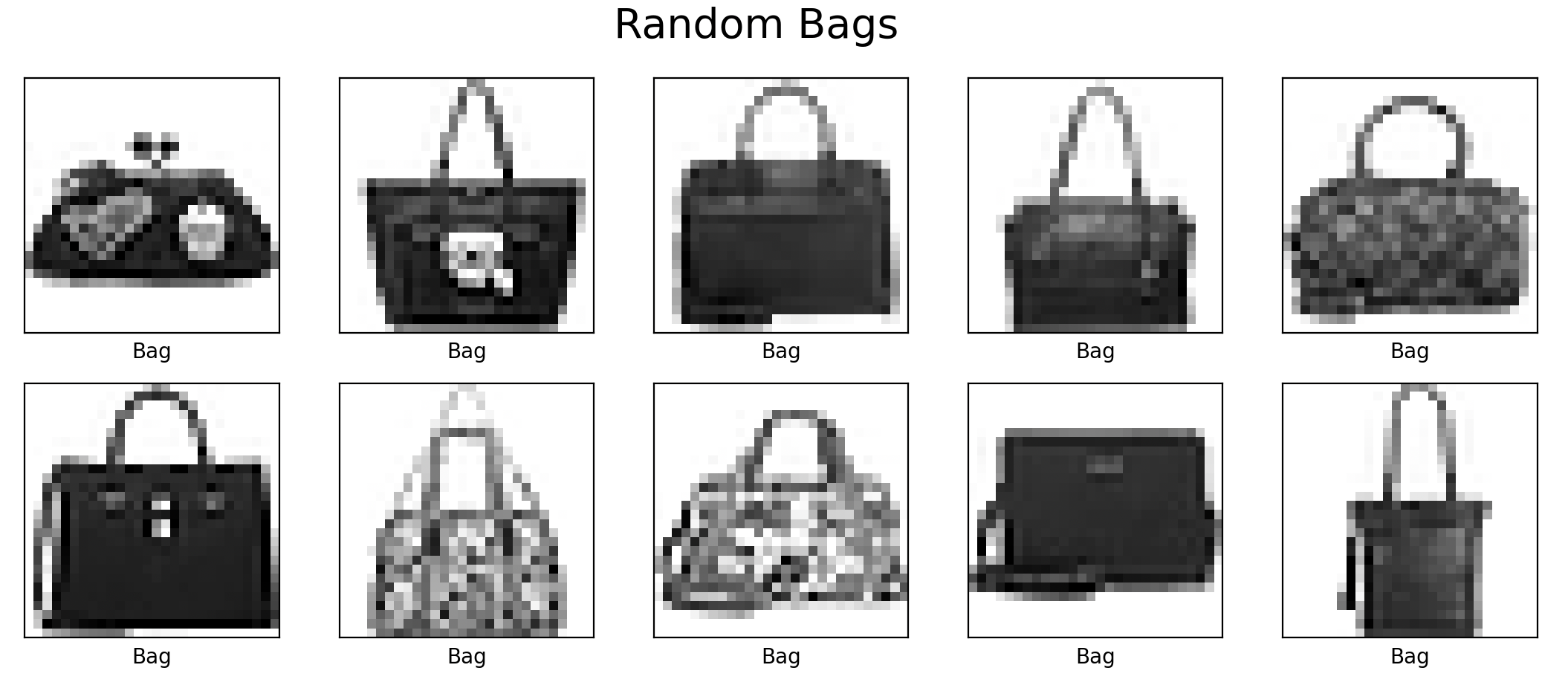} }}
    \caption{Left: Chosen "Bag" images of \ourmethodabbrev, Right: Chosen "Bag" images of Random. Rectangular shaped bags form the core of the full bag class (by visual inspection). The bags chosen by \ourmethodabbrev\ are consistent and similar, and are all  rectangular in shape and similar sizes. Random bags, however, also include a variety of more uncommon outlier shapes, including a purse and long, thin bag. Under a data constraint, it is more important to get the core rectangular bag correct, as \ourmethodabbrev\ chooses.}
    \label{fig:bag-examples}
\end{figure}

\paragraph{2D Embeddings.} First, we perform a tSNE dimensionality reduction on the random set and \ourmethodabbrev\ set in \figref{fig:tsne}. The diagrams show that \ourmethodabbrev\ selects more core, densely clustered data and leads to better separability between classes. It does not choose much outlier data for each class that is spread around the space and could confuse the model. Random, on the other hand, is generally less separable and more sparse, and includes potentially confusing points in each class that are spread around the space and overlapping with other classes, which could confuse the model. In short, \ourmethodabbrev-selected data is more dense and separable than random and is likely to lead to better classification performance, but will have higher KL divergence from full data than the random data because of this effect on the distribution.

\paragraph{Visualization.} As an additional sanity check, we visually inspect the images chosen by \ourmethodabbrev\ and random images in \figref{fig:bag-examples}. We show the bag class as an example. \ourmethodabbrev\ selects more consistent bags that conform to each other. They are all rectangular and similar in size, and represent the core set of bag images, as regular rectangular bags form most of the class (as determined by visualizing samples from the full set). Random images, on the other hand, contain more strange images like triangular bags and tote bags and are less consistent. Under a data constraint, \ourmethodabbrev\ focuses on the core rectangular bags that best represent the class and discards outliers, which leads to better performance. 

\paragraph{Separability.} Finally, we examine the separability between classes as an indicator of whether \ourmethodabbrev\ selects images that are more core and less spread/confusing. We compute the Euclidean mean for each class and compute the minimum distance between that mean to the means of all other classes, for \ourmethodabbrev-selected data and random images. \tabref{tab:distances} shows the results, and demonstrates that \ourmethodabbrev\ classes are further away from each other in 8 out of the 10 classes. This implies \ourmethodabbrev\ selects more distinct and less confusing points, and likely explains the improved classification performance over random images.

\begin{table}[tp]
  \caption{Minimum Euclidean distance between the mean of each class and the closest other class (i.e. closest neighbor). \textbf{Higher} is better, as it indicates the class is further apart from others and therefore more separable. \ourmethodabbrev\ is higher in 8 out of the 10 classes, indicating \ourmethodabbrev-selected data is more separable than random images.}
  \label{tab:distances}
  \centering
  {\begin{tabular}{l c c}
    \toprule
    \textbf{Class} & \textbf{\ourmethodabbrev} & \textbf{Random}\\
    \midrule
    0 & 922 & \textbf{962}\\
    1 & \textbf{1123} & 1044\\
    2 & \textbf{604} & 600\\
    3 & \textbf{1123} & 1044\\
    4 & \textbf{604} & 600\\
    5 & \textbf{1102} & 1005\\
    6 & \textbf{723} & 664\\
    7 & \textbf{1102} & 1005\\
    8 & 1269 & \textbf{1361}\\
    9 & \textbf{1660} & 1570\\
    \bottomrule
  \end{tabular}}
\end{table}


\subsection{Label Distribution}
One of the core assumptions of the method is that we can utilize only the input side of data without consideration for labels. By reconstructing an approximate $X$ using $G$ with only the input side, we implicitly reconstruct the label side of $X$ as well, and therefore there is no need to consider the labels explicitly. We verify this assumption with FashionMNIST, as it has distinct, quantifiable labels unlike the text generation tasks. \tabref{tab:label-distribution} shows the distribution of labels recovered by \ourmethodabbrev, and shows that we approximately recover an even 10\% for each class, which is the distribution of the full $X$. Therefore, \ourmethodabbrev\ recovers the distribution of labels in $X$ implicitly, despite not including them in the algorithm, as desired.
This is not to say labels are not important in data selection. We note that for the algorithm to work well, there should exist a quality relation between input and label. Specifically, we require that either input and label are good quality, or input and label are bad quality. If a good quality input maps to a wrong or bad quality label, we may recover bad labels. 

\begin{table}[tp]
  \caption{Distribution of labels per class. Despite only considering inputs, \ourmethodabbrev\ also recovers the label distribution of the target $X$.}
  \label{tab:label-distribution}
  \centering
  \resizebox{\textwidth}{!}{\begin{tabular}{l c c c c c c c c c c}
    \toprule
    \textbf{Class} & 0 & 1 & 2 & 3 & 4 & 5 & 6 & 7 & 8 & 9\\
    \midrule
    \textbf{\% of Total} & 8\% & 10.1\% & 10.2\% & 13.3\% & 10\% & 8.5\% & 10.9\% & 7.1\% & 11.2\% & 10.6\%\\
    \bottomrule
  \end{tabular}}
\end{table}

\begin{figure}[tp]
\centering
\includegraphics[width=\textwidth]{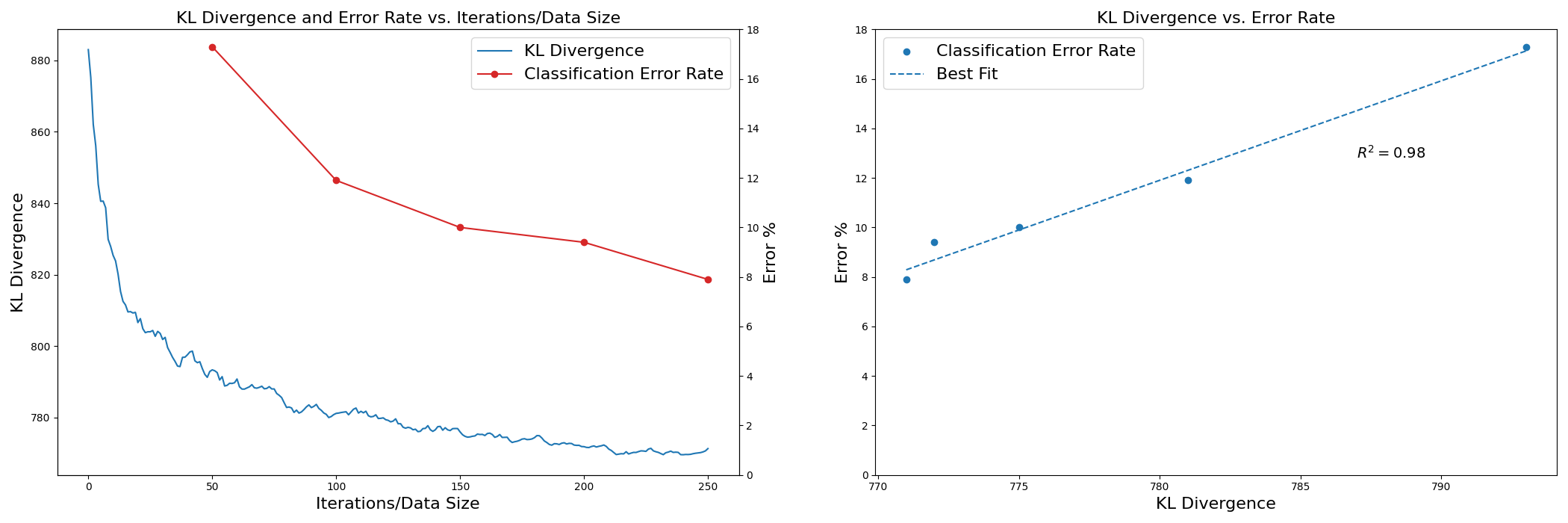}
\caption{Left: KL divergence and Classification Error Rate over different data sizes. The shape of the two curves are very similar. Right: KL divergence and Classification Error Rate. High $R^2=0.98$ indicates Kl divergence can accurately predict performance.}
\label{fig:kl-div-err-rate}
\end{figure}

\begin{table}[tp]
  \caption{Performance of the model using the different datasizes chosen as the algorithm progresses.}
  \label{tab:data-size-performances}
  \centering
  \setlength{\tabcolsep}{12pt}
  {\begin{tabular}{l c c r}
    \toprule
    \textbf{Iterations/Data \%} & \textbf{Size Train/Valid} & \textbf{Accuracy} & $\mathbf{\hat D_{KL}}$\\
    \midrule
    50/5\% & 3,000/1700 & 82.7\% & 793\\
    100/10\% & 6,000/1700 & 88.1\% & 781\\
    150/15\% & 9,000/1700 & 90.0\% & 775\\
    200/20\% & 12,000/1700 & 90.6\% & 772\\
    250/25\% & 15,000/1,700 & 92.1\% & 771\\
    \bottomrule
  \end{tabular}}
\end{table}

\subsection{Correlation Between Performance and KL Divergence}

We know that under the FashionMNIST problem setup for reducing training set size, a data size was chosen beforehand, which in practice would represent a budget. However, beyond budget concerns, we endeavored to find out if we could use the KL divergence to predict what a good, convergent data size would be. Perhaps 25\% is more than enough and we could do just as well with less, or perhaps 25\% is too little to get good performance, and we might want to increase our resources before experimenting. In this setup, we know the KL divergence values at each data size (each iteration), and we want to see whether the shape of the KL divergence curve could indicate the shape of the performance curve at the different data sizes. We repeat the same experiment described in \secref{sec:fashionmnist}, but using 50 iterations (5\% data), 100 iterations (10\% data), 150  iterations (15\% data) and 200 iterations (20\% data). To keep the validation set at 1700, if there is not enough data left after selecting the right train size, we supplement the set with training data points from the general set. We plot the results of both KL divergence and performance (classification error rate) in \figref{fig:kl-div-err-rate} and describe the results in \tabref{tab:data-size-performances}. In \figref{fig:kl-div-err-rate}, we additionally plot error rate as a function of KL divergence and compute the coefficient of determination $R^2$. We find that the shape of classification error vs. data size and KL divergence vs. data size are very similar. Additionally, $R^2=0.98$, which indicates a high degree of correlation between KL divergence with data size and performance. Theoretically, this implies that we can use the KL divergence to pick an ideal data size if we are under budget constraints: where the KL divergence curve flattens, we are unlikely to receive much gain in performance from increasing the data size further and can use that data size with the expectation of decent performance. We only have a few datapoints to support this hypothesis, however, and we leave exploration of this property of the algorithm to future work.
\clearpage
\section{Guidance for Choosing X, D, Algorithm Parameters}

We provide a brief note on choosing X, D and various algorithm parameters.

\subsection{X}
The target set must be chosen carefully, as this is the distribution the algorithm will match to. If it is misrepresentative or biased, the chosen train set will be as well. For systems where the model may be used in production whose values are known (e.g. on a search engine's queries), it makes sense to pick X as a representative sample of that production traffic and align model training distribution to that. If the algorithm is being used to select high quality, aligned, or unbiased data, it makes sense for $X$ to be a carefully curated set of human-annotated data. 

It is equally important to make $X$ as broad as possible within the desired target. If $X$ misses a portion of the space (e.g. misses long search queries, for example), then the chosen data will also not capture that area of the space and the resulting model will be prone to poor generalization.

\subsection{D}
The initialization set is versatile. It can be a set that is known to be high quality, like human-annotated data, or it can be initialized as a random subset of $G$. Generally, initializing $D$ as a random subset of $G$ will give better results when the data size is smaller overall (e.g. our EN-DE WMT experiment), but matters less when the data size is large (e.g our EN-FR WMT experiment).

\subsection{Algorithm}
We have covered in \ref{app:algorithm} many considerations for various aspects of the algorithm, in particular the stopping criterion. In general, we recommend trying different parameters and observing the intial convergence of the KL Divergence, as this is cheap to do once the K Means has taken place (we recommend saving those so it doesn't need to be recomputed). For some tasks and data (e.g. WMT) the convergence is smooth and strict criterion is good enough. For other tasks (e.g. FashionMNIST) the convergence has some ups and downs and it serves to choose another criterion like max\_sequential\_increases or simply terminating the algorithm when the KL divergence bottoms out. If the algorithm diverges, we recommend increasing the uniform start size as well as decreasing the learning rate if it is set high.

For the v\_init parameter, "mean" will reset the algorithm to the mean of $X$ to start gradient descent every time. We recommend setting a higher learning rate with this parameter, as this enables the algorithm to reach different points in the space (since it starts from the same point every time). This value is good for when the target distribution may be tightly clustered together (e.g. WMT). Similarly, "prev\_opt" will start from the previous optimal point and is suitable for tightly clustered distributions where we expect the new optimal point not to be far from the previous one. We recommend a medium learning rate for this value. On the other hand, using the "jump" value is more suitable for exploring the space and adding diversity, particularly for multimodal or spread-out distributions (e.g. FashionMNIST). We recommend a medium learning rate with this value as well.

As for resets like in the spelling correction experiments, we recommend enabling this when it is believed that having multiple copies of the same data point is useful for learning. This may be when we want to weight certain datapoints or regions of the space more (which will let the algorithm choose on it's own how best to weight the space).


    




\end{document}

%% file: algorithm.tex
\begin{algorithm}[tp]
\caption{\ourmethod}\label{alg:cap}
\begin{spacing}{1.2}
\begin{algorithmic}

\State{\textbf{Quantize} $X,D,G$ using K-means and pick the cluster centroids $X_c, D_c,G_c$ as the new points}
\While{\textit{Not Stopping Criterion}}
\State{\textbf{Gradient-Descend to find $\mathbf{v}_{opt}$:}}

\State \hskip1.5em $\mathbf{v}_1 \gets $ previous $\mathbf{v}_{\textit{opt}}$, $\bar{\mathbf{x}}$ or random \Comment{We explore different techniques in our experiments}
\State \hskip1.5em \textbf{Perform} $ \mathbf{v}_{k+1} \leftarrow \mathbf{v}_k-\gamma \cdot \frac{\partial}{\partial \mathbf{v}_k} \hat D_{\textit{KL}}(P_{X_c} \parallel P_{D_c \cup \{\mathbf{v}_k\}})$ until converged to $\mathbf{v}_{\textit{opt}}$
\State{\textbf{Update $D_c$ :}}
\State \hskip1.5em $\mathbf{v}_b \gets \argmin_{\mathbf{v}_i \in G_c} ||\mathbf{v}_i-\mathbf{v}_{\textit{opt}}||$ \Comment{The closest point in $G_c$ to $\mathbf{v}_{\textit{opt}}$}

\State \hskip1.5em $D_c \gets D_c + \{\mathbf{v}_b\}$
\State \hskip1.5em \textbf{Remove} $\mathbf{v}_b$ from $G_c$

\EndWhile

\State{\textbf{Explode:} Select points from full $D$ and $G$ which belong to the chosen centroids’ ($D_c \cup V_c$) clusters}
\end{algorithmic}
\end{spacing}
\end{algorithm}

%% file: explosion.tex
\begin{figure}[tp]
\centering
\includegraphics[width=0.85\textwidth]{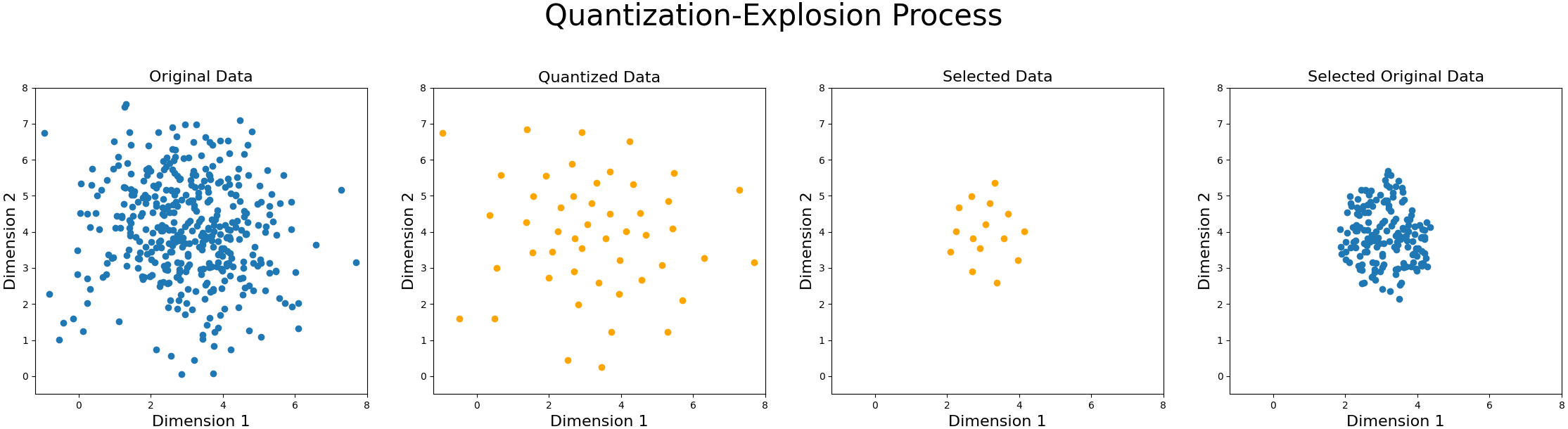}
\caption{Visualization of the Quantization-Explosion Process. From left to right: original data (400 points), representative K-means centroids (50 points) of the original data (Quantization), selected centroids after data selection, original data represented by the selected centroids (Explosion)}
\label{fig:quantization-explosion}
\end{figure}

%% file: consistency.tex
\begin{figure}[tp]
\centering
\includegraphics[width=0.85\textwidth]{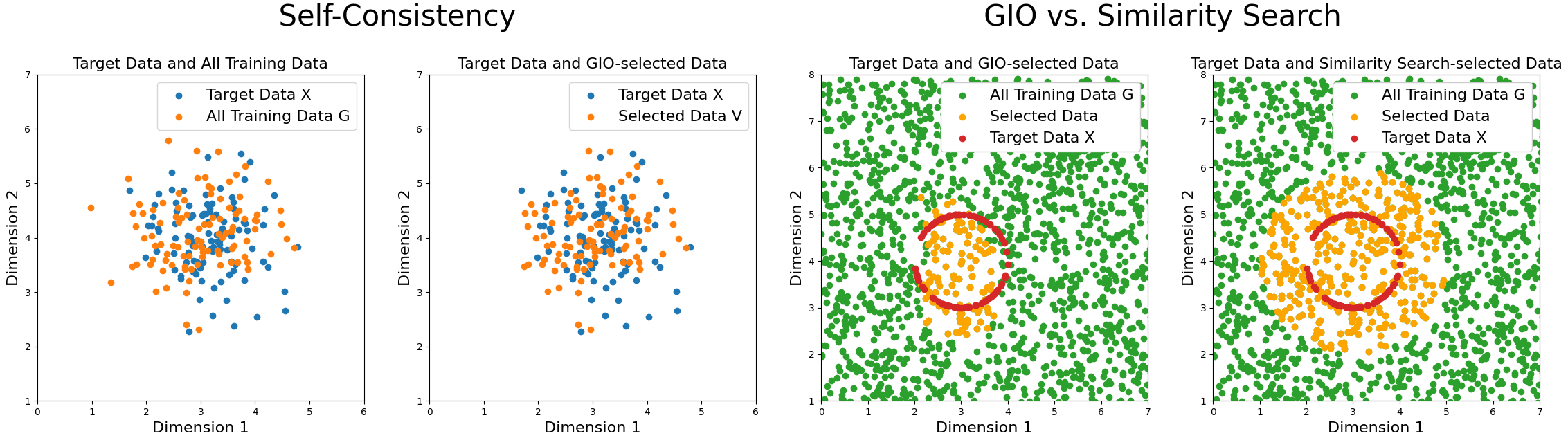}
\caption{The leftmost graph shows $X$ and $G$, which come from the same distribution. The second graph shows that \ourmethodabbrev\ recovers nearly all of $G$ (self consistency). The right two graphs compare \ourmethodabbrev\ with similarity search. Points within the circle formed by $X$ are more ideal than points outside. By considering the distribution, \ourmethodabbrev\ selects nearly all points inside before terminating (third graph). By comparison, in order to pick points within the circle, similarity search also picks a range of points outside the circle, which is suboptimal (fourth graph).}
\label{fig:consistency}
\end{figure}